\documentclass[journal]{IEEEtran}
\usepackage{amsmath,amsfonts,amssymb,latexsym,epic,epsfig,psfrag,booktabs}
\usepackage{multirow}
\usepackage{subfigure}
\usepackage{float}
\begin{document}

\title{Automated Steel Bar Counting and Center Localization with Convolutional Neural Networks}
%
%
%

\author{Zhun Fan,~\IEEEmembership{Senior Member,~IEEE,}
        Jiewei Lu,
        Benzhang Qiu,
        Tao Jiang,
        Kang An,
        Alex Noel Josephraj,
        and Chuliang Wei

\thanks{Corresponding author: Chuliang Wei (email: clwei@stu.edu.cn).}
\thanks{Jiewei Lu and Benzhang Qiu contributed equally.}
\thanks{Authors are with the Guangdong Provincial Key Laboratory of Digital Signal and Image Processing, College of Engineering, Shantou University, Shantou 515063, China.}
\thanks{This research work was supported by Guangdong Key Laboratory of Digital Signal and Image Processing, the National Natural Science Foundation of China under Grant (61175073, 61300159, 61332002, 51375287).}}





\maketitle

\begin{abstract}
  Automated steel bar counting and center localization plays an important role in the factory automation of steel bars. 
  Traditional methods only focus on steel bar counting and their performances are often limited by complex industrial environments. 
  Convolutional neural network (CNN), which has great capability to deal with complex tasks in challenging environments, is applied in this work. 
  A framework called CNN-DC is proposed to achieve automated steel bar counting and center localization simultaneously. 
  The proposed framework CNN-DC first detects the candidate center points with a deep CNN. 
  Then an effective clustering algorithm named as Distance Clustering (DC) is proposed to cluster the candidate center points and locate the true centers of steel bars. 
  The proposed CNN-DC can achieve 99.26\% accuracy for steel bar counting and 4.1\% center offset for center localization on the established steel bar dataset, 
  which demonstrates that the proposed CNN-DC can perform well on automated steel bar counting and center localization. 
  Code is made publicly available at: https://github.com/BenzhangQiu/Steel-bar-Detection.
\end{abstract}

\begin{IEEEkeywords}
  Steel bar counting, center localization, convolutional neural network, distance clustering.
\end{IEEEkeywords}

%
\IEEEpeerreviewmaketitle

\section{Introduction}
\IEEEPARstart{S}{teel} industry is one of the most important basic industries for many countries. 
Steel bar is one of the most commonly used steel products, which is mainly used for building construction. 
Factory automation \cite{aawagner1990power,bbmiyatake1993experiencing,cackorber2007modular,dadgrau2017industrial,eaedotoli2017advanced,fafdotoli2018overview} plays an essential role in improving the productivity of steel bars, 
in which automated steel bar counting and center localization are among the most crucial steps.

\subsection{Steel Bar Counting}
Steel bar counting plays an important role in the management of steel production. 
Traditional steel bar counting is based on human calculation. 
Skilled workers count the number of steel bars in the factory, which is time-consuming and prone to errors. 
In recent years, some image processing techniques are employed to achieve automated steel bar counting. 
In \cite{gzhang2008bar}, Zhang $et.al.$ use a template matching algorithm and a mutative threshold segmentation method to achieve steel bar counting. 
In \cite{hying2010research}, Ying $et.al.$ combine Sobel operator and Hough transformation for automatic steel bar counting. 
In \cite{iwu2015steel}, Wu $et.al.$ propose a steel bar counting method which utilizes concave dots matching, K-level fault tolerance and visual feedback. 
In \cite{jghazali2017automatic}, Ghazali $et.al.$ employ Hough transformation and the Laplacian of Gaussian (LoG) technique to perform automated steel bar  counting. 
In \cite{kxiaohu2018research}, Liu $et.al.$ propose a contours-based steel bar identification algorithm to count the number of steel bars. 
However, these methods are easily affected by the noisy and complex industrial environment, and need to adjust many parameters during operations.

\subsection{Center Localization}
\begin{figure}
  \centering
\includegraphics[width=3.0in]{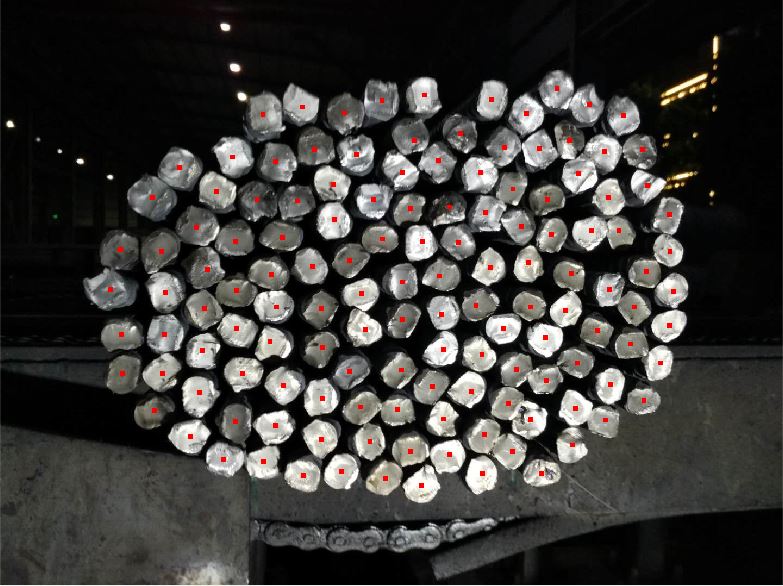}
  \caption{An example of center localization. Red points are the centers of steel bars.}
\label{fig:centerLo}
\end{figure}

\begin{figure}
  \centering
\includegraphics[width=3.0in]{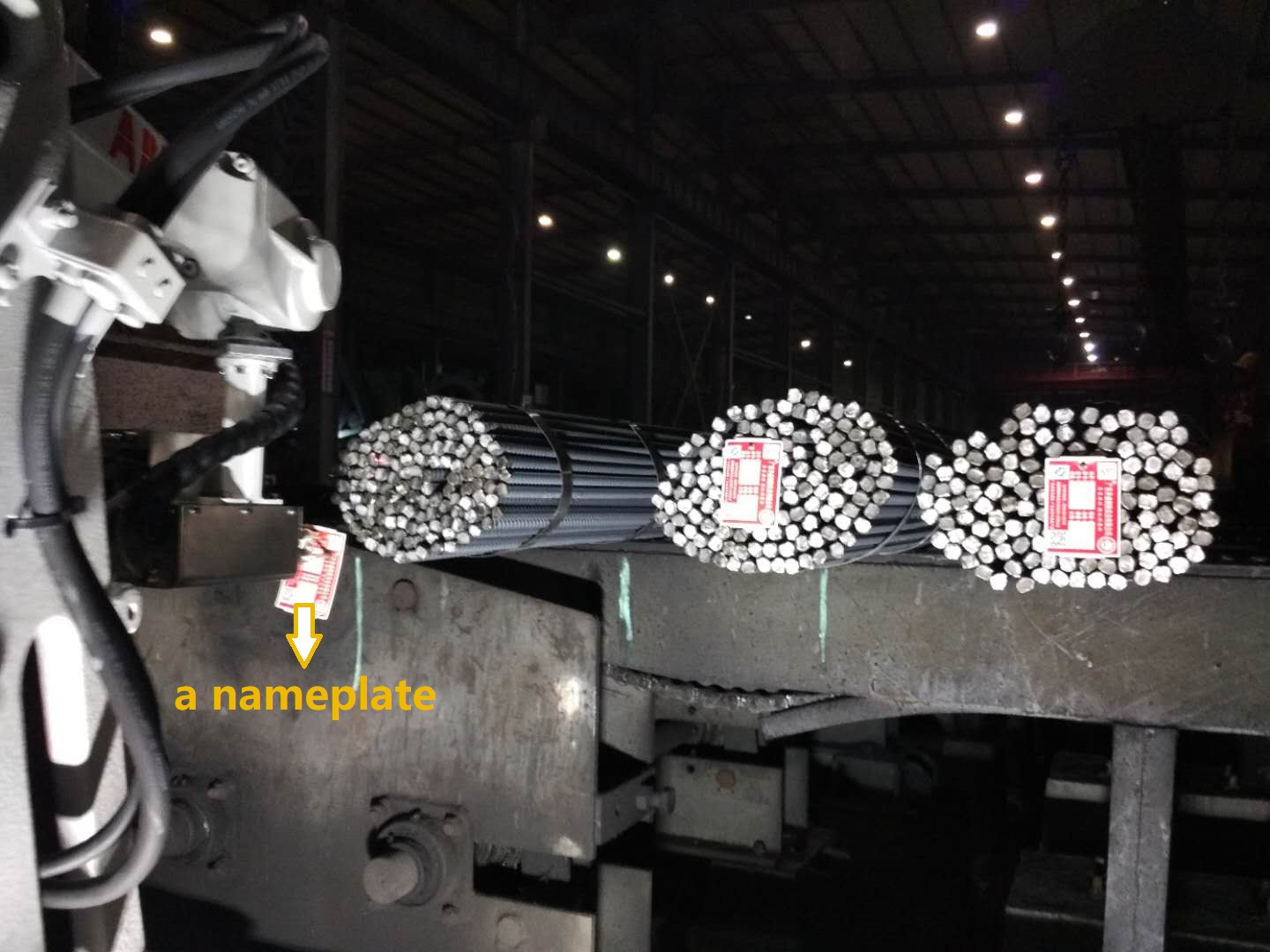}
  \caption{Manipulators are used to weld nameplates on the steel bars.}
\label{fig:centerLo1}
\end{figure}

Center localization means locating the center of each steel bar, as shown in Fig. \ref{fig:centerLo}. 
The reason of performing center localization is that it is an important step to achieve automated nameplate welding. 
Before leaving steel factories, steel bars need to be welded nameplates which contain the information of steel bars, such as production time and type specification. 
Traditional nameplate welding is performed by skilled workers. 
In the process of automated production, manipulators are used to weld nameplates on the steel bars, as shown in Fig. \ref{fig:centerLo1}. 
Before a nameplate on a steel bar is welded, an appropriate welding point which is close to the center of a steel bar needs to be chosen in order to avoid the broken and sliding problems of nameplates in transit. 
Therefore, locating the centers of steel bars is the first and one of the most crucial steps to achieve automated nameplate welding. 

\begin{figure*}
  \centering
\includegraphics[width=6.5in]{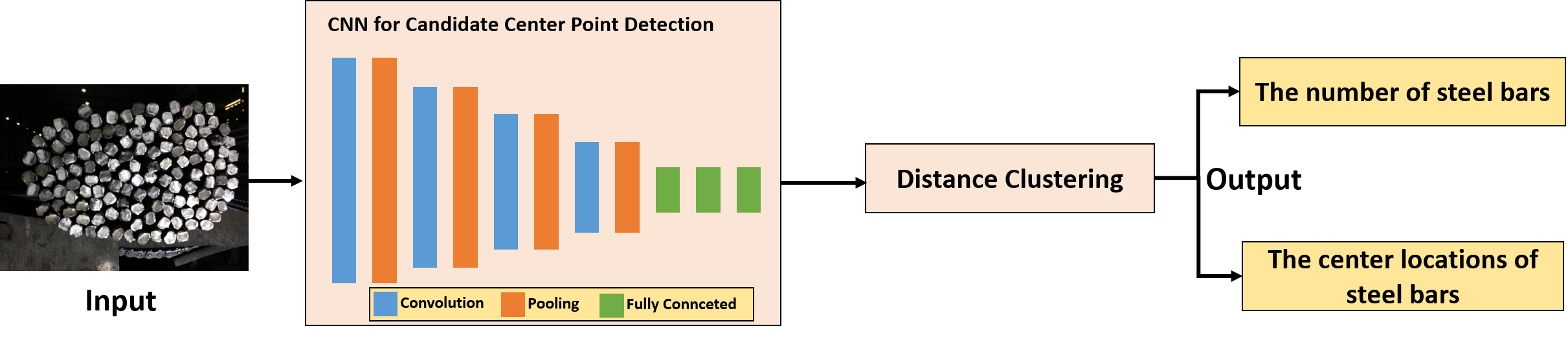}
  \caption{The proposed CNN-DC framework. CNN-DC first detects the center points with a deep CNN, and then applies Distance Clustering algorithm to obtain the information of the number of steel bars and the center locations of the steel bars. }
\label{fig:flowchart}
\end{figure*}

In this paper, a framework called CNN-DC is proposed to achieve automated steel bar counting and center localization simultaneously. 
The proposed CNN-DC framework first detects the candidate center points with a deep convolutional neural network (CNN). 
Then an effective clustering algorithm called \emph{Distance Clustering (DC)} is proposed to cluster the candidate center points and obtain the centers of steel bars.  
The experimental results demonstrate the effectiveness of CNN-DC on steel bar counting and center localization.

The rest of this paper is structured as follows: Section II describes the proposed CNN-DC framework. Section III introduces the steel bar dataset and evaluation metrics. Section IV provides the experimental results. Section V presents the conclusions of this paper.

\section{METHODOLOGY}
The proposed CNN-DC framework can be regarded as a two-stage algorithm for automated steel bar counting and center localization. 
CNN-DC first obtains the candidate center points of steel bars with a deep CNN, and then an effective clustering algorithm named as Distance Clustering is proposed to cluster the candidate center points. 
Fig. \ref{fig:flowchart} shows the framework of CNN-DC.

\subsection{Convolutional Neural Networks}
Convolutional neural networks are applied in our work to detect the candidate center points. 
Neural networks are inspired by biological processes \cite{lmatsugu2003subject,mlecun1998gradient}, and can be used to process a variety of high dimensional data \cite{nabdel2014convolutional, acharya2018deep, phu2014convolutional,qren2017faster, rsu2015multi, ssun2015learning, sakurai2019restoring, zheng2017wide, gao2018object}, such as images, videos, voice signals and text characters. 
When dealing with the abovementioned data, the application of fully connected networks is sometimes cumbersome due to its large feature space. 
Therefore, CNNs as special types of neural networks are preferred due to some important characteristics, such as spatial arrangement, sparse interactions, parameter sharing \cite{krizhevsky2012imagenet}.

The input of CNNs is called tensor, which comprises of a multi-dimensional array. 
The core components of CNNs are convolutional and pooling layers. 
A convolutional layer convolves the input tensor with a set of kernels to generate the output tensor. 
For each kernel, a feature map is generated by performing convolution, which slides the kernel on the whole spatial positions of the input tensor. 
Each convolutional layer consists of a set of kernels and thus produces a collection of feature maps, which are stacked together to generate the output tensor. 
When the input tensor is shaped as greyscale image, the following steps are performed in order to specify the convolutional layer: 
(1) Accepting the input tensor with size $H_i\times W_i \times D_i$, and obtaining the four parameters: $K$(the number of kernels), $F$ (the spatial dimensions of kernels), $S$(stride) and $P$(the zero padding size). 
(2) Producing the output tensor with size $H_o\times W_o \times D_o$, where $H_o=(H_i-F+2P)/S+1$, $W_o=(W_i-F+2P)/S+1$ and $D_o=K$. 
(3) The total number of parameters for the kernels is $(F\times F \times D_i) \times K$, where each kernel has $(F\times F \times D_i)$ parameters. 
Pooling layers are used to reduce the number of training parameters and control overfitting. 
They have two downsampling strategies: max-pooling and mean-pooling. 
Normally max-pooling is adopted with kernels of size $2\times 2$ and stride $2$.

\begin{figure*}
  \centering
\includegraphics[width=6.5in,height=2in]{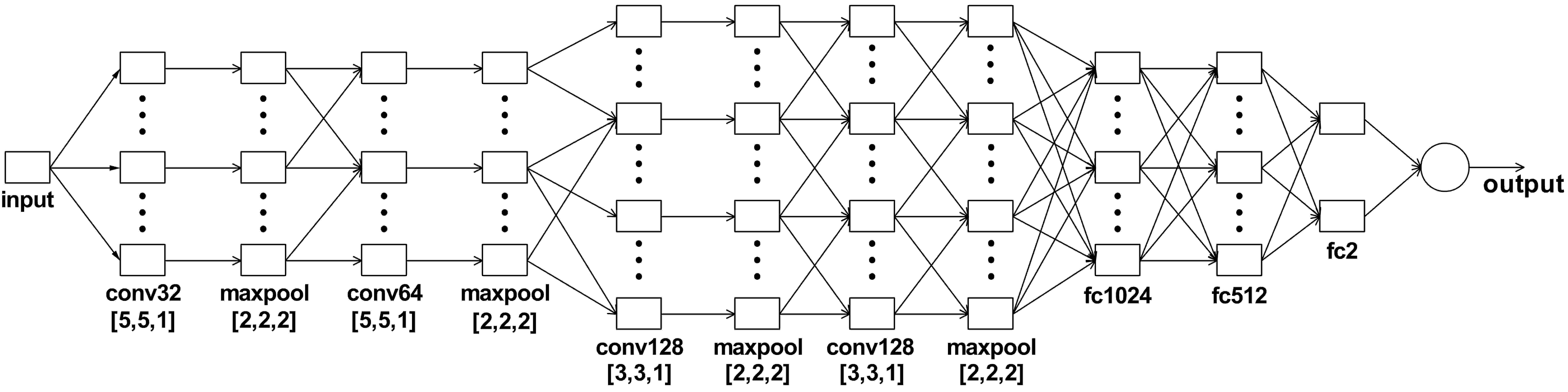}
  \caption{The CNN architecture composed of four convolutional layers, four pooling layers and three fully connected layers. Layer names are followed by the number of feature maps. Square brackets specify the kernel size and stride. It is noted that 'conv', 'maxpool' and 'fc' are short terms for convolutional layer, max pooling layer and fully connected layer, respectively. Zero-padding is not used in this paper.  }
\label{fig:arch}
\end{figure*}

Convolutional neural networks generally consist of several convolutional layers and pooling layers, finalized with one or more fully connected layers. 
The details of CNNs used in this paper are provided in Fig. \ref{fig:arch}.

\subsection{Distance Clustering}
Clustering is a statistical analysis method applied to classification problems. The methods of clustering include: connectivity-based clustering \cite{zhang1996birch,karypis1999chameleon}, 
centroid-based clustering \cite{hartigan1979algorithm,ng2002clarans}, 
density-based clustering \cite{ester1996density} \cite{rodriguez2014clustering},
grid-based clustering \cite{ankerst1999optics}.
An effective clustering algorithm called Distance Clustering (DC) is proposed to cluster the candidate center points obtained from the CNN.

\begin{figure}
  \centering
\includegraphics[width=3.4in]{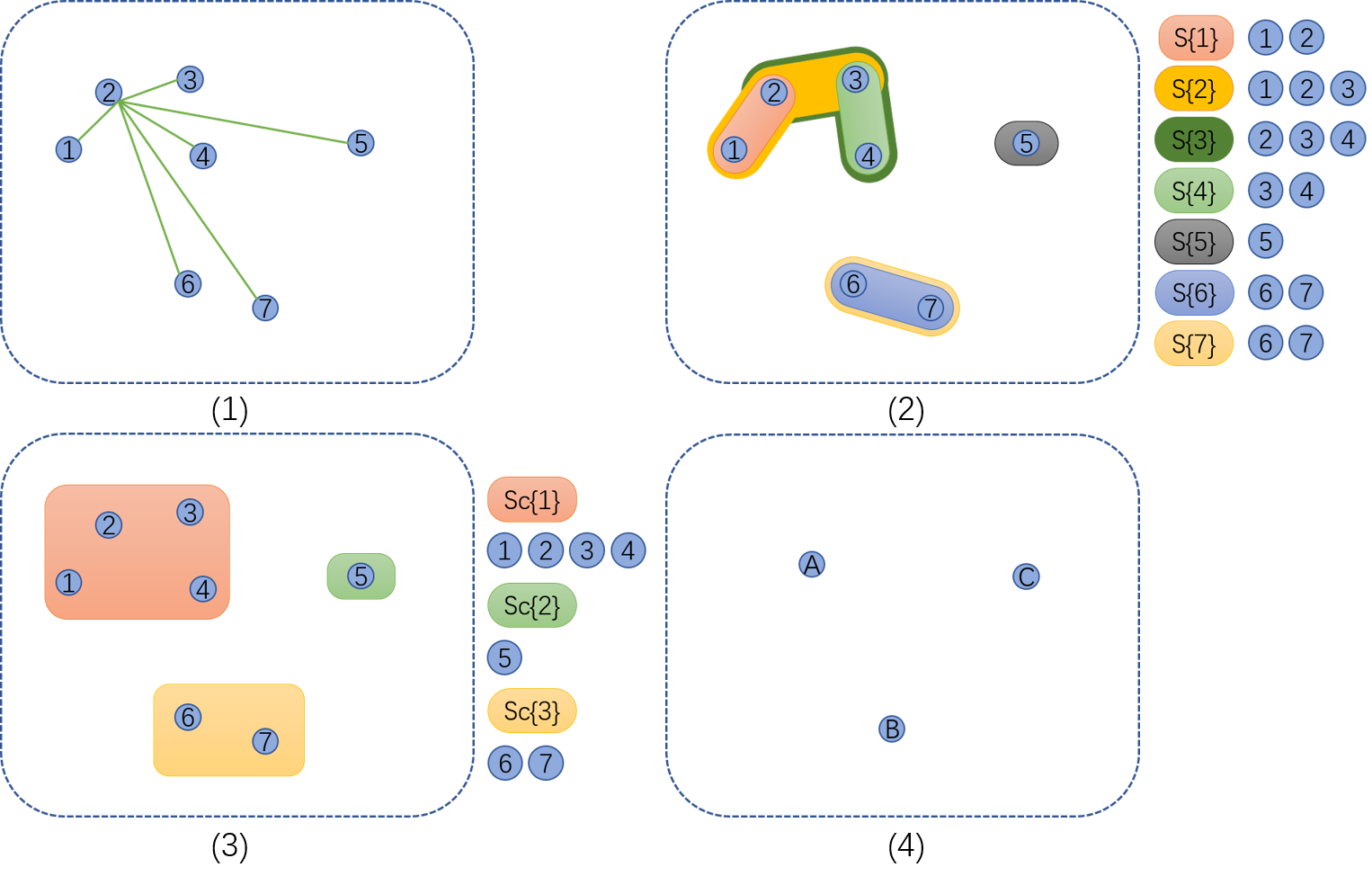}
  \caption{Illustrating the process of distance clustering: 
  (1) Calculating distances of each two candidate center points. 
  (2) Creating a set for each candidate center point by distance threshold. 
  (3) Merging sets if they have common elements.
  (4) Calculating the center point of each set.}
\label{fig:clusterflow}
\end{figure}

The pseudocode of DC is shown in Algorithm 1 and illustrated in Fig. \ref{fig:clusterflow}:
\vspace{1mm}

\scalebox{0.98}{
\begin{tabular}{p{8cm}}
\hline
\textbf{Algorithm 1}: Distance Clustering (DC) \\
\hline
  \textbf{Input:} The locations of candidate center points and a distance threshold $th_d$  \\
  \textbf{Output:} A set of clusters, the center of each cluster and the number of clusters\\
  \textbf{Step 1:Initialization}\\
  1) For $i = 1, \dots, n$, set $D(i)=d_{ij}$, where $n$ is the number of candidate center points, $d_i$ is the Euclidean distance between the $i$th candidate center point and its closest candidate center point $j$, $D$ is the set of $d_{ij}$.    \\
  2) Create an empty structure $S$ to save the initial clusters. \\
  3) For $i = 1, \dots, n$, do \\
   \quad \quad   if D(i) $<th_d$, S\{i\} = \{i;j\}; else S\{i\} = \{i\} \\
   \quad  end for \\

   \textbf{Step 2:Clustering}\\
  1) Create a structure $S_c\{1\}=S\{1\}$ to save the final clusters and set the number of clusters as $n_{S_c}$ = 1. \\
  2) For $i=2,\dots,n$,do         \\
  \quad \quad $n_{S_c}$ = length($S_c$)\\
  \quad \quad condition = 0\\
  \quad \quad  For $k=1,\dots,n_{S_c}$,do \\
 \quad \quad \quad if $(S\{i\}\cap S_c\{k\})!=\varnothing$, do \\
  \quad \quad \quad \quad $S_c\{k\} = \{S_c\{k\};S\{i\}\}$                           \\
  \quad \quad \quad \quad                        condition = 1 \\
  \quad \quad \quad \quad                        break \\
  \quad \quad \quad end if \\
  \quad \quad end for \\
  \quad \quad if condition = 0, do \\
  \quad \quad \quad $n_{S_c}=n_{S_c}+1$ \\
  \quad \quad \quad $S_c\{n_{S_c}+1\} = S\{i\}\}$ \\
  \quad \quad end if \\
  \quad  end for \\
  \textbf{Step 3:Obtaining Centers}\\
  1) Create an empty center set $C$ to save the centers \\
  2) For $i=1,\dots,n_{S_c}$,do         \\
  \quad \quad  $x_{c_i}$ = (max($x_{S_c\{i\}}$) + min($x_{S_c\{i\}}$)) / 2\\
  \quad \quad $y_{c_i}$ = (max($y_{S_c\{i\}}$) + min($y_{S_c\{i\}}$)) / 2\\
  \quad  \quad $C(i)$ = $(x_{c_i},y_{c_i})$   \\
   \quad end for \\
\hline
\end{tabular}}
\vspace{1mm}

\emph{\textbf{Initialization:}} 
In this step, for the $i$th candidate center point, its Euclidean distances with other candidate center points are calculated first. 
Then the closest distance $d_{ij}$ with the $j$th candidate center point is chosen and assigned to the $i$th candidate center point. 
Moreover, a structure $S$ is created to save neighbors for the $i$th candidate center point. 
A distance threshold $th_d$ is used to decide whether the point $j$ is a neighbor of the $i$th candidate center point and belongs to the set of $S\{i\}$. 
In the experiment, $th_d$ is set as 20.

\emph{\textbf{Clustering:}} A new structure $S_c$ is created to save the final clusters. For each clusters $S\{i\}$ in $S$, if it has common elements with one of the clusters $S_c\{k\}$  in $S_c$, the cluster $S\{i\}$ will be merged into $S_c\{k\}$. Otherwise, $S_c$ will create a new cluster to save $S\{i\}$.

\emph{\textbf{Obtaining Centers:}} For each cluster in $S_c$, the localization of each cluster center is calculated by averaging the maximum and minimum coordinate values for the x-coordinate and y-coordinate.

\section{Dataset and Evaluation Metrics}
\subsection{Dataset}
The steel bar dataset consists of 10 images. The images are obtained by an industrial camera in a steel bar factory. 
Each image has a high resolution and is of size $1440\times 1080$. 
Moreover, each image has RGB channels and each channel is with 8 bits. 
The 10 images were divided into a training set and test set. 
The training set contains 4 images while the test set consists of 6 images.
99195 patches extracted from the training set are used to train the network.

\subsection{Evaluation Metrics}
Four commonly used evaluation metrics are applied in our work to assess the performance of CNN-DC:
\begin{eqnarray*}
      && Recall = \frac{TP}{TP+FN}   \\
      && Precision = \frac{TP}{TP+FP}   \\
      && F1 = \frac{2\times Precision \times Recall}{Precision + Recall}
\end{eqnarray*}

where TP, FP and FN indicates true positive (the number of correctly detected center points), false positive (the number of incorrectly detected center points), and false negative (the number of undetected center points), respectively. 
$Recall$ indicates the CNN-DC's ability of detecting center points while $Precision$ is used to measure the CNN-DC's capability of correctly detecting center points.
 $F1$ is a comprehensive index of $Recall$ and $Precision$. The calculation time of applying CNN-DC on each test image is also stored.


In order to further evaluate the performance of CNN-DC for automated steel bar counting and center localization, two other useful metrics are employed in this work.

The first one is relative accuracy \cite{uarmstrong1992error} :
\begin{equation}\label{equ:error}
  Acc_r = (1-\frac{|N_d-N|}{N})\times 100\%
\end{equation}
where $N_d=TP+FP$ is the number of detected center points. $N$ is the actual number of steel bars. $Acc_r$ evaluates the performance of CNN-DC on steel bar counting.

The second is the offset degree of center points:
\begin{equation}\label{equ:error}
  offset = \frac{\Sigma_{i=1}^{N}\frac{X_i}{m}}{N}\times 100\%
\end{equation}
where $m=71$ is the average diameter of a steel bar and used to perform data normalization, $X_i$ is the Euclidean distance between the $i$th manually marked center point and the closest detected center point. $offset$ is used to evaluate the performance of CNN-DC on center localization.

\begin{figure}
  \centering
\includegraphics[width=2.7in]{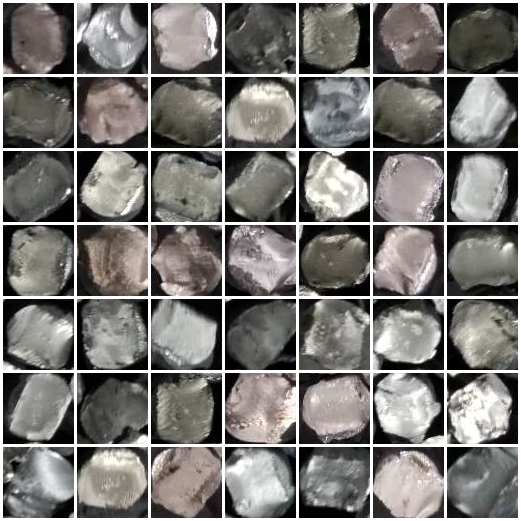}
  \caption{The patches with label 1.}
\label{fig:1p}
\end{figure}

\begin{figure}
  \centering
\includegraphics[width=2.7in]{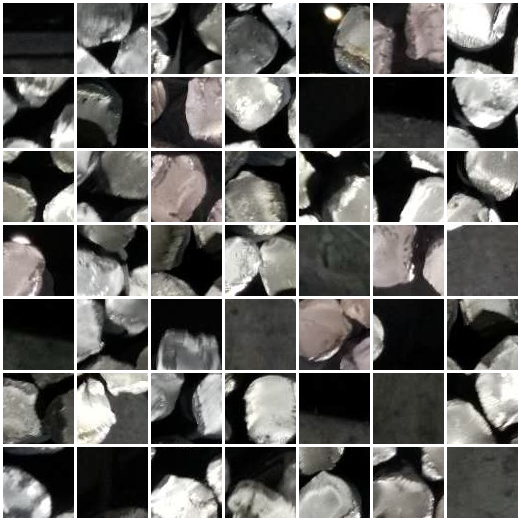}
  \caption{The patches with label 0.}
\label{fig:0p}
\end{figure}


\section{Experiments}

\begin{figure*}
  \centering
\includegraphics[width=7in]{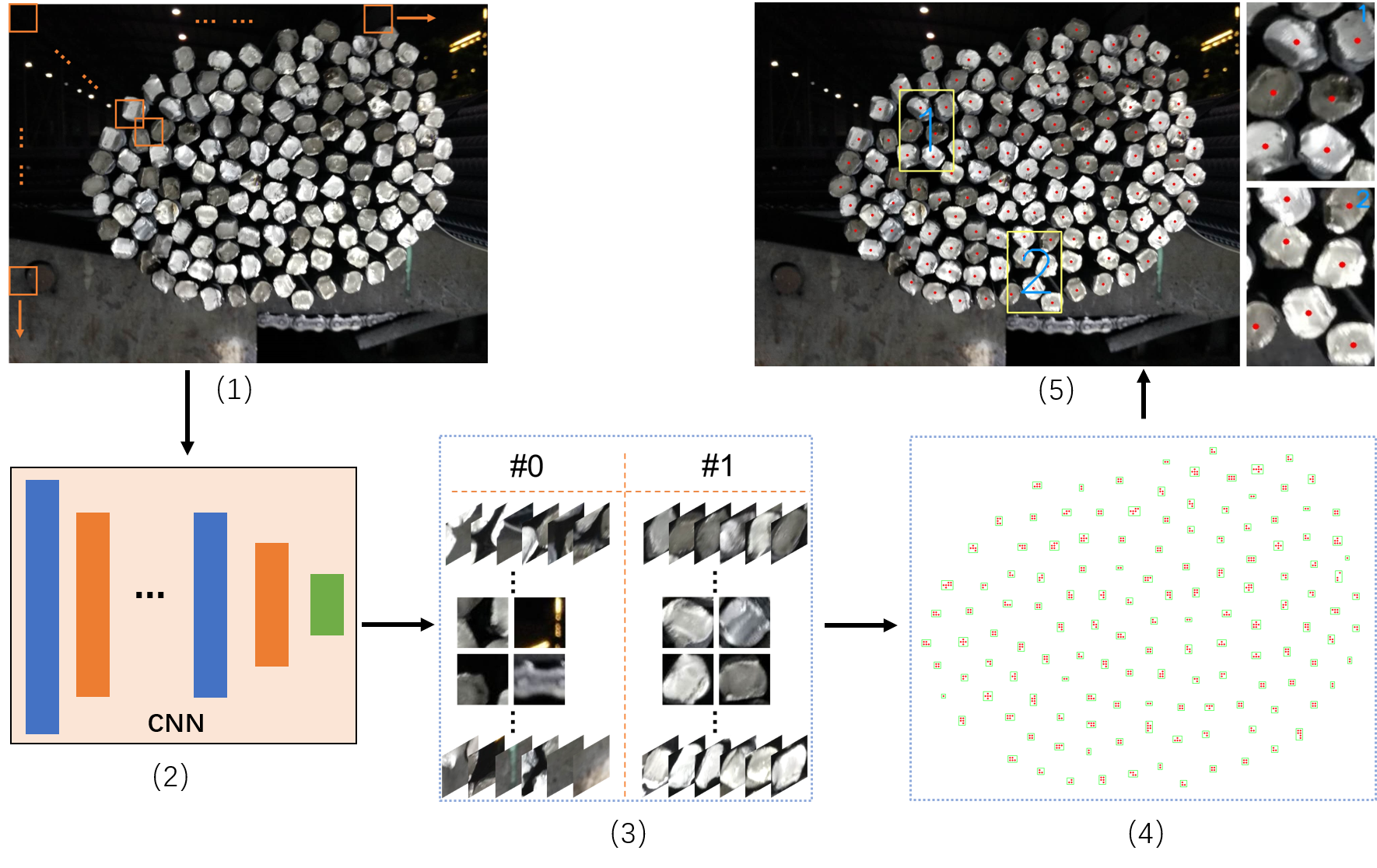}
  \caption{The process illustrating the application of CNN-DC: 
  (1) A $71\times 71$ sliding window is adopted to extracted patches from images based on each image pixel.
  (2) These patches are detected by CNN with architecture shown in Figure \ref{fig:arch}.
  (3) The result of CNN classification. 
  (4) The center coordinates of the patches considered as label 1 are shown as red points, and the green boxes show the group of red points by distance clustering with process shown in Algorithm 1. 
  (5) The center of each group is reimposed into the center of each steel bar of the original image accordingly. }
\label{fig_flow}
\end{figure*}

In this section, the training setup of CNN is first introduced, followed by the experimental results.
\subsection{Training Setup}
\subsubsection{Network Architecture and Training Parameters}
The network architecture used in our experiment is composed of four convolutional layers, four pooling layers and three fully connected layers, as shown in Fig. \ref{fig:arch}. 
The network was trained by the stochastic gradient descent algorithm \cite{vbottou2012stochastic}. $L_2$ regularization with a weight decay 0.0001 was adopted to prevent overfitting. 
The learning rate was set as 0.001 and the training was stopped after 40 epochs. The implementation of CNN-DC was based on Tensorflow \cite{wkbadi2016tensorflow}. 
The training was conducted on a Intel Xeon E5-2690 CPU with a TITAN Xp GPU.



\subsubsection{Data Preparation}
In our experiment, 4 steel bar images are used to train the network. 
The input of the network are patches extracted from images based on each image pixel. The patch size is of $71\times 71$ ($71$ is approximately the diameter of the steel bars). 
$1440\times1080\times4$ patches are extracted first. Then the patches whose centers were within the $7\times7$ rectangles centered in the manual center points are labeled as 1, and other patches are labeled as 0. 
Finally, 26468 patches are labeled as 1, while 6194332 patches are labeled as 0, which leads to an imbalanced data problem. 
In training, all patches with label 1 are chosen first. 
Then the patches with label 0 are selected randomly according to the ratio of positive (1) to negative (0) patches. 
In our experiment, the ratio of positive to negative patches is set as $1:3$. 
Thus 26468 patches with label 1 and 72727 patches with label 0 are used to train the network. 
The examples of positive and negative training examples are shown in Fig. \ref{fig:1p} and \ref{fig:0p}. 
During testing, a sliding window with stride $6$ was employed to improve the efficiency of CNN-DC. 
The process of selecting the optimal test stride with regard to the efficiency of CNN-DC is provided in the third experiment of the next section.

\begin{table}[htbp]
  \centering
  \caption{The performance of CNN-DC on the steel bar dataset}
    \begin{tabular}{cc|ccc}
    \hline
    \hline
    \multicolumn{2}{c|}{\multirow{4}[4]{*}{Average Indexs of CNN-DC}} & Recall & Precision & F1 \\
    \multicolumn{2}{c|}{} & 0.9951  & 0.9976  & 0.9963  \\
\cline{3-5}    \multicolumn{2}{c|}{} & $Acc_r$ & $offset$ & times(s) \\
    \multicolumn{2}{c|}{} & 99.26\%  & 4.11\%  & 3.5862  \\
    \hline
    \hline
    \end{tabular}%
  \label{tab:ex1}%
\end{table}%

\begin{figure*}
  \begin{tabular}{cc}
  \begin{minipage}{0.15\linewidth}
  \begin{center} Zhang $et.al.$\cite{gzhang2008bar}\end{center} 
  \end{minipage}
  \begin{minipage}{0.42\linewidth}
  \includegraphics[width=2.8in]{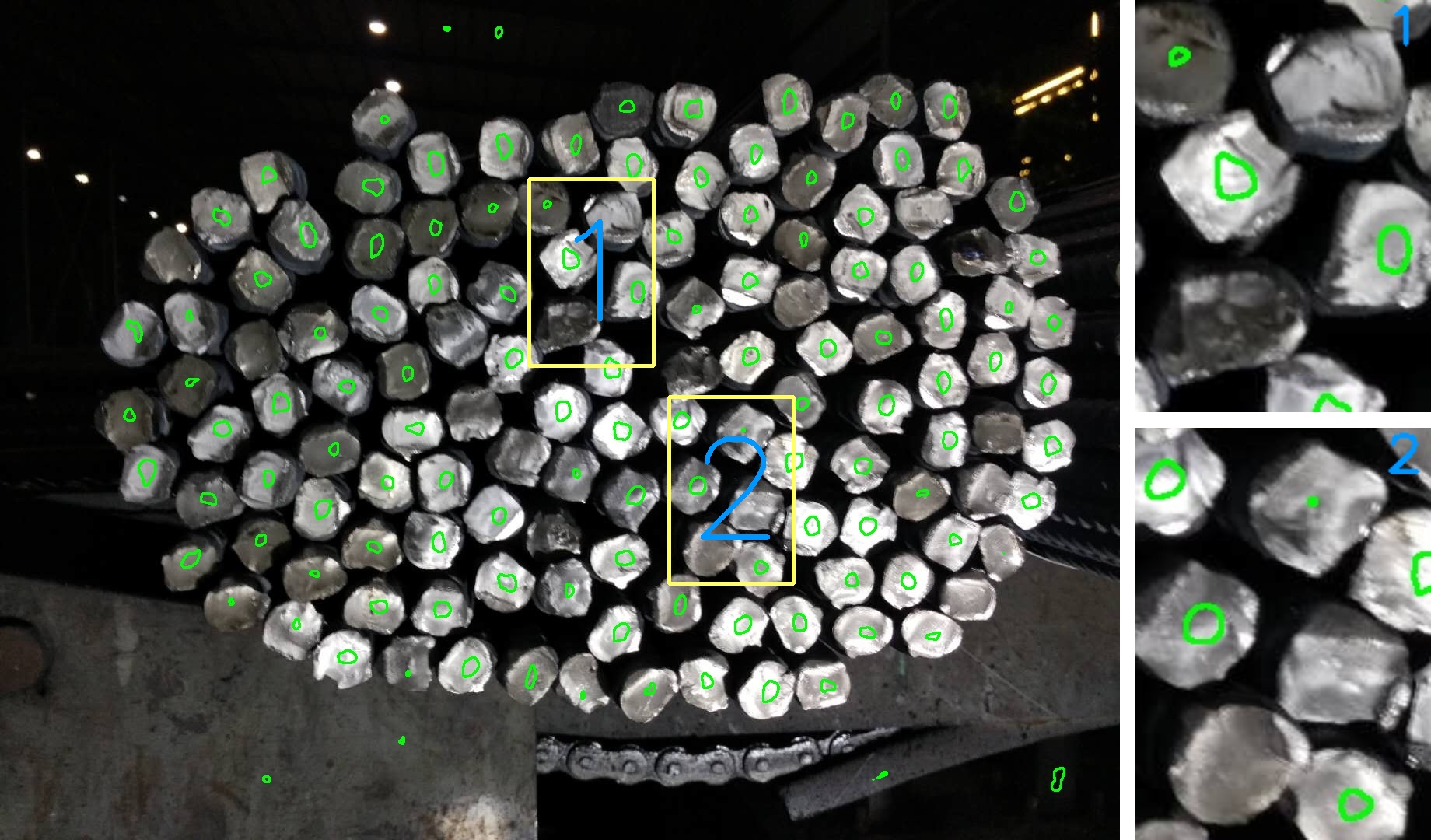}
  \end{minipage}
  \begin{minipage}{0.42\linewidth}
  \includegraphics[width=2.8in]{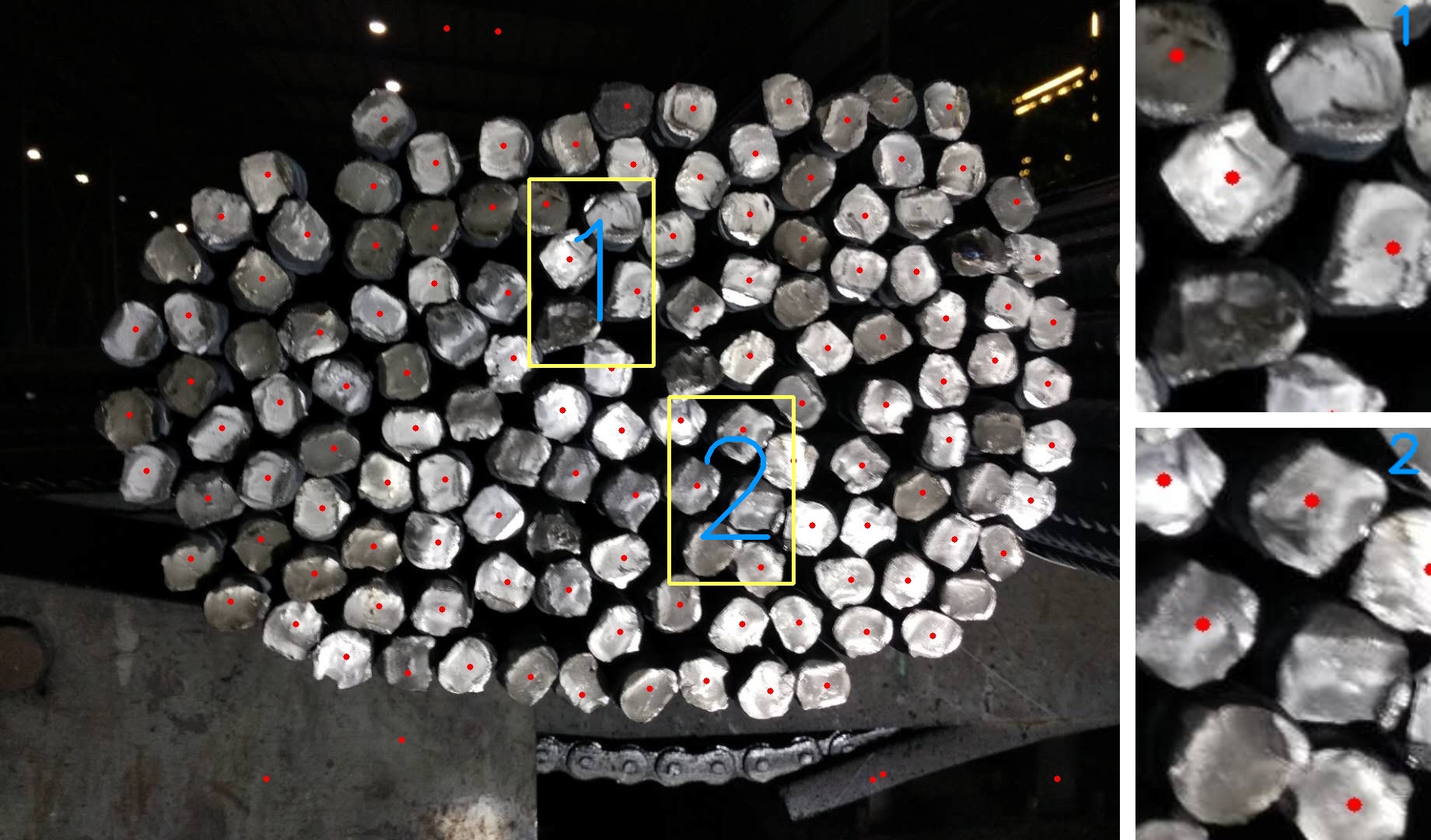}
  \end{minipage}
  \vspace{1mm}
  \end{tabular}

  \begin{tabular}{cc}
    \begin{minipage}{0.15\linewidth}
    \begin{center} Ying $et.al.$\cite{hying2010research}\end{center} 
    \end{minipage}
    \begin{minipage}{0.42\linewidth}
    \includegraphics[width=2.8in]{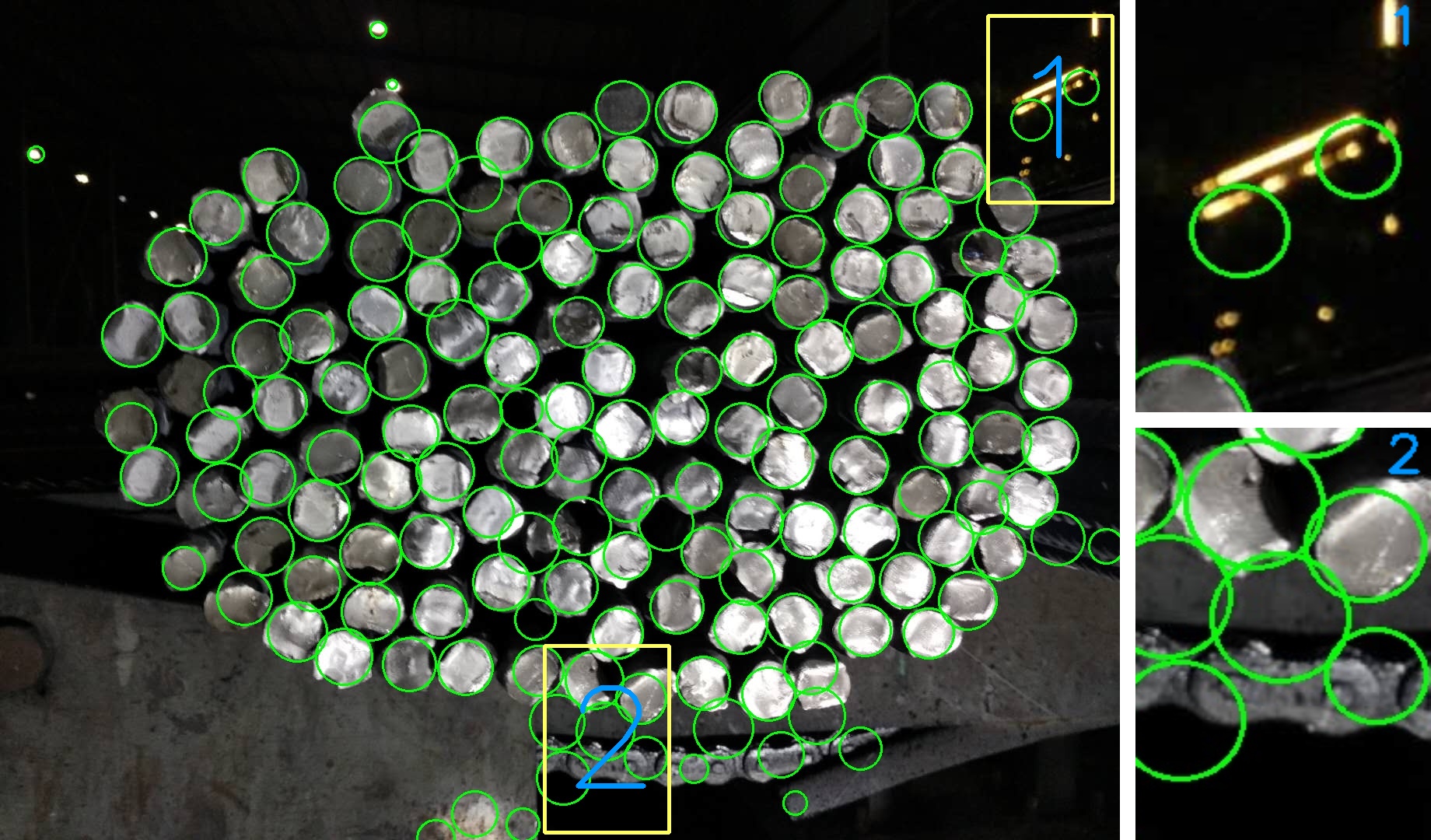}
    \end{minipage}
    \begin{minipage}{0.42\linewidth}
    \includegraphics[width=2.8in]{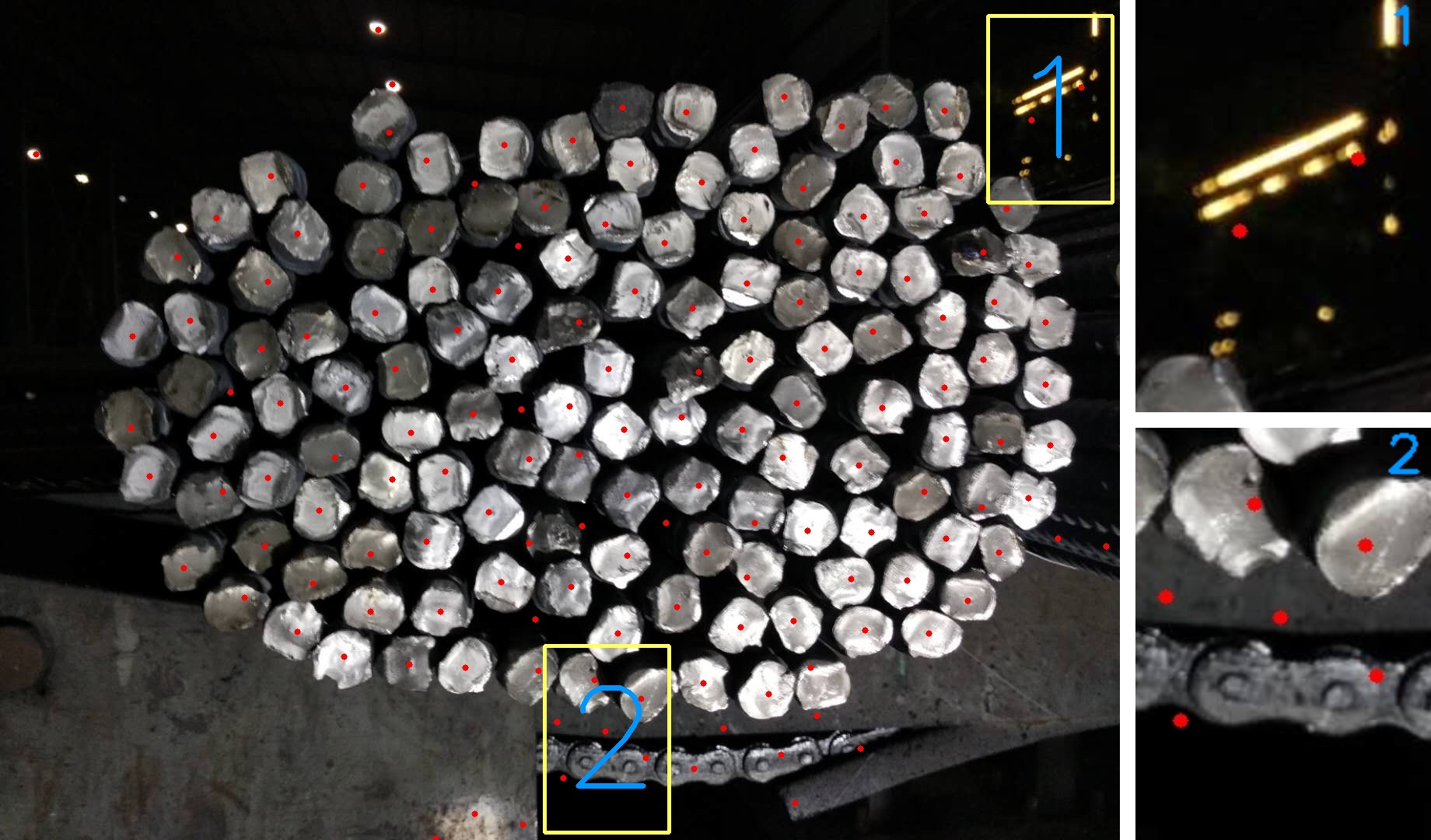}
    \end{minipage}
    \vspace{1mm}
    \end{tabular}
  
  \begin{tabular}{cc}
    \begin{minipage}{0.15\linewidth}
    \begin{center} Ghazali $et.al.$\cite{jghazali2017automatic}\end{center} 
    \end{minipage}
    \begin{minipage}{0.42\linewidth}
    \includegraphics[width=2.8in]{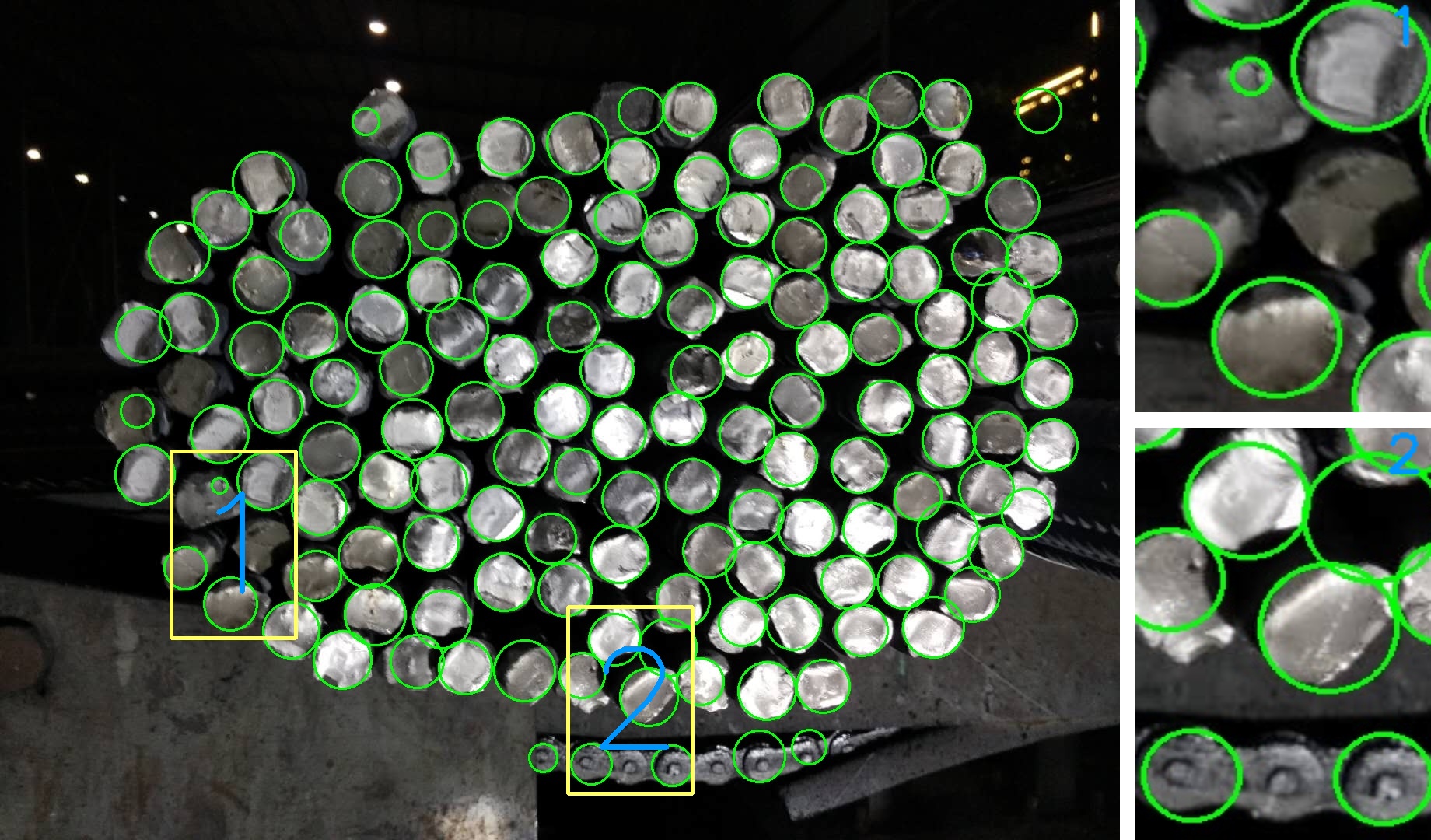}
    \end{minipage}
    \begin{minipage}{0.42\linewidth}
    \includegraphics[width=2.8in]{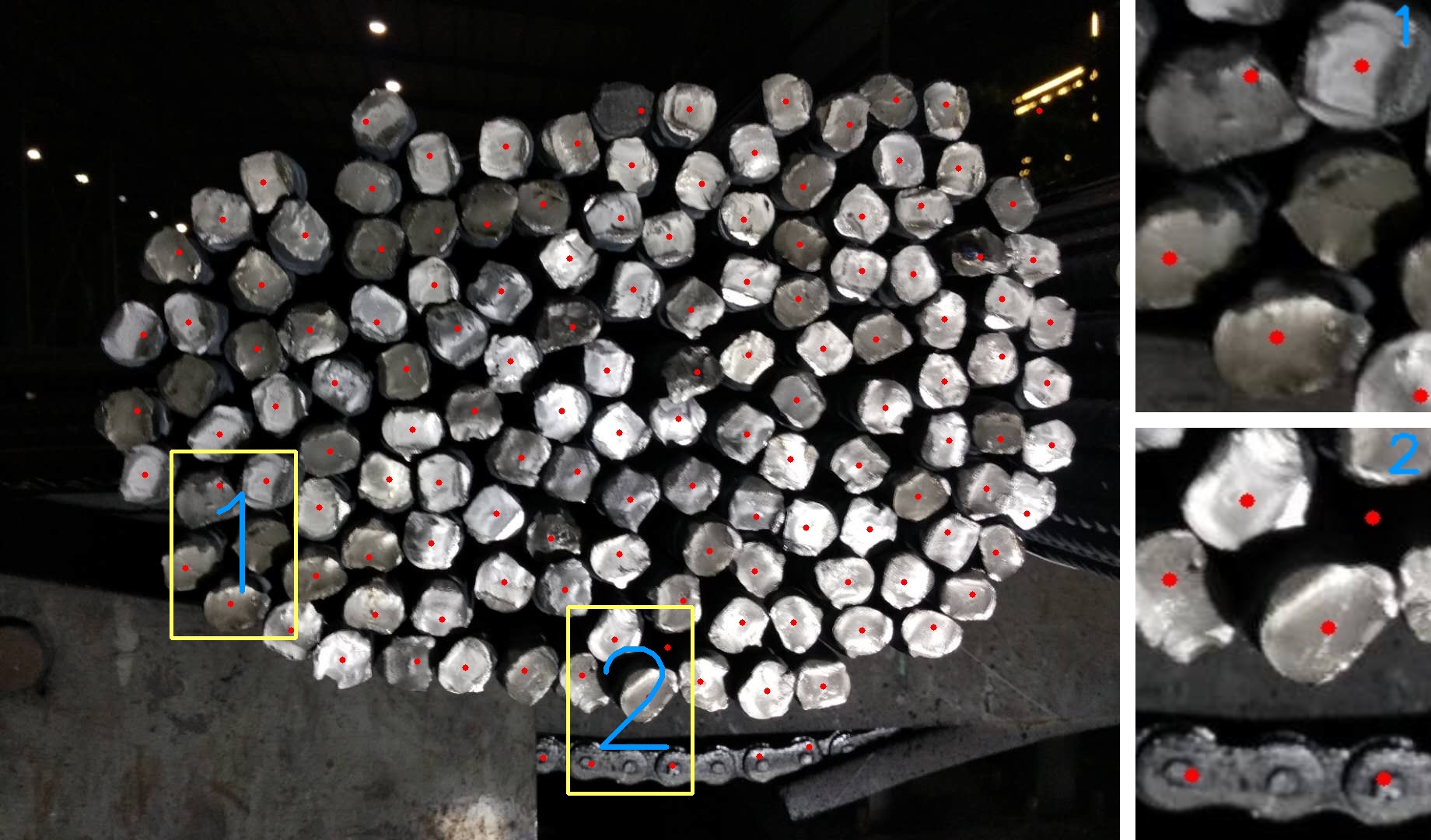}
    \end{minipage}
    \vspace{1mm}
    \end{tabular}
  
  \begin{tabular}{cc}
    \begin{minipage}{0.15\linewidth}
    \begin{center} Liu $et.al.$\cite{kxiaohu2018research}\end{center} 
    \end{minipage}
    \begin{minipage}{0.42\linewidth}
    \includegraphics[width=2.8in]{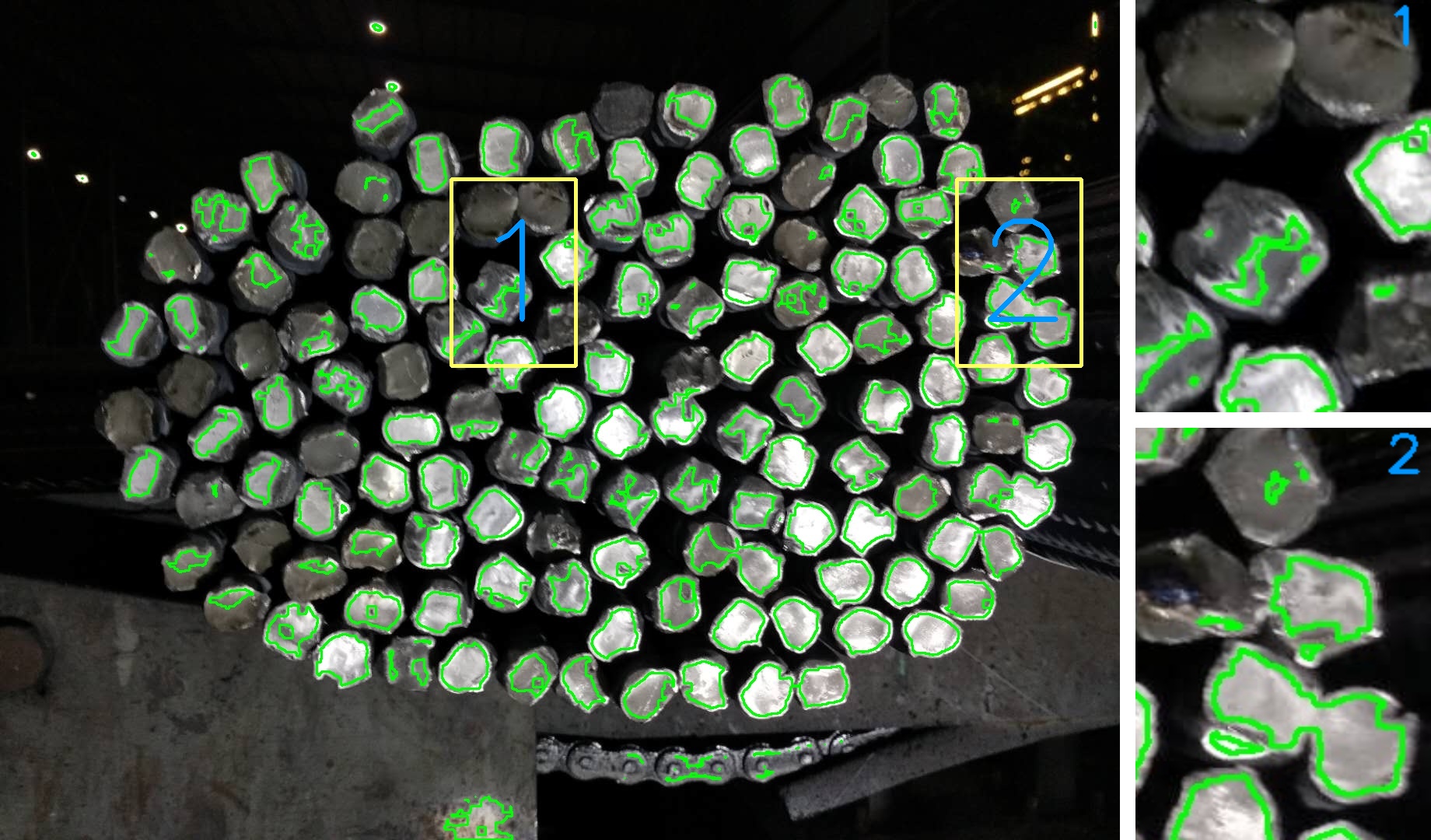}
    \end{minipage}
    \begin{minipage}{0.42\linewidth}
    \includegraphics[width=2.8in]{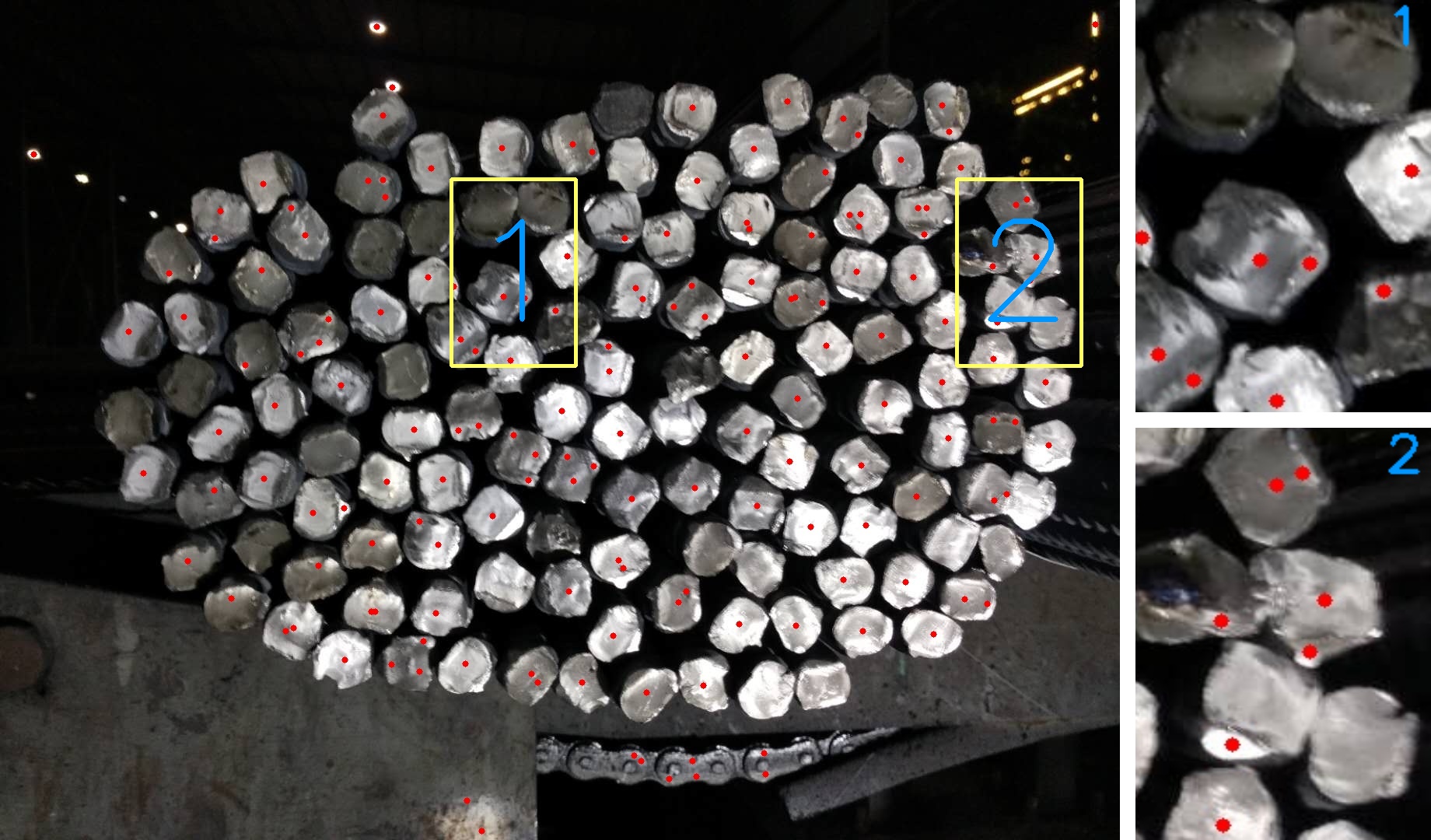}
    \end{minipage}
    \vspace{1mm}
    \end{tabular}
  
  \begin{tabular}{cc}
    \begin{minipage}{0.15\linewidth}
    \begin{center} \textbf{Proposed}\end{center} 
    \end{minipage}
    \begin{minipage}{0.42\linewidth}
    \includegraphics[width=2.8in]{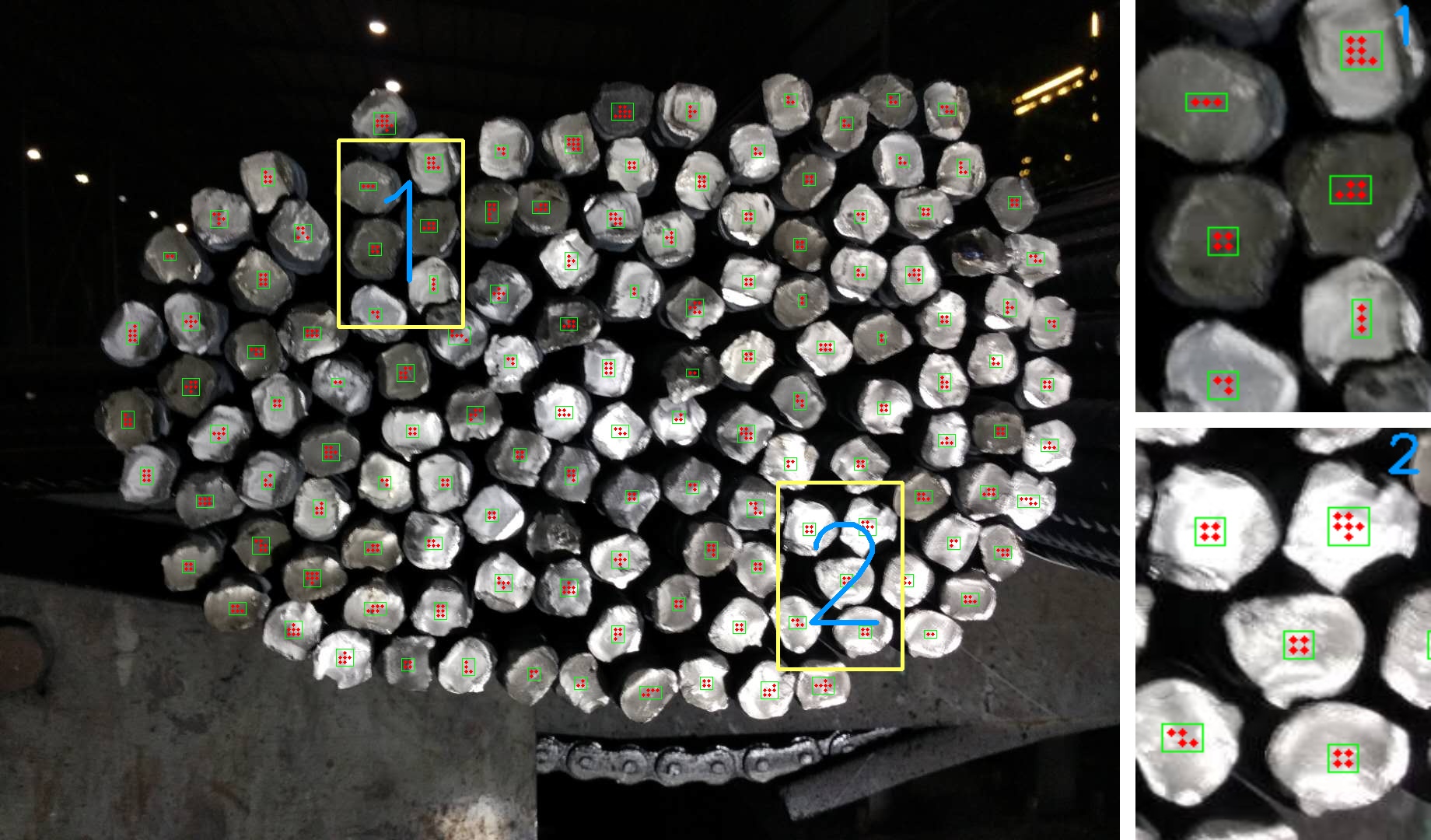}
    \end{minipage}
    \begin{minipage}{0.42\linewidth}
    \includegraphics[width=2.8in]{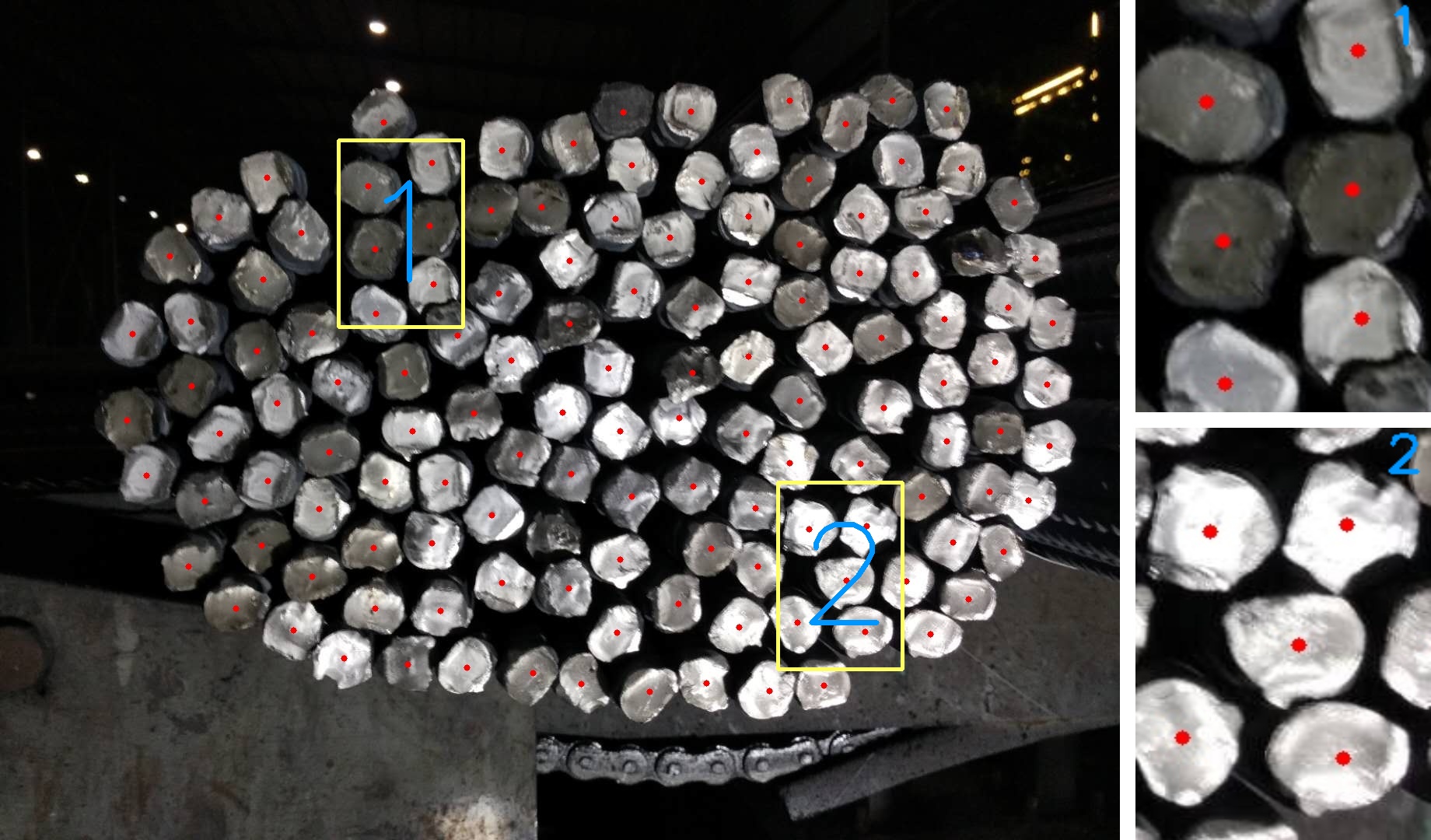}
    \end{minipage}
    \end{tabular}
    \center{\hspace{2cm}The intermediate result \hspace{5cm} The final result}
\caption{Comparison of results obtained by different methods on an exemplar image.
In Zhang $et.al.$\cite{gzhang2008bar}, the template matching algorithm can not match the image which is not similar to the template, and fails to identify some steel bars.
In Ying $et.al.$\cite{hying2010research} and Ghazali $et.al.$\cite{jghazali2017automatic}, the Hough transformation algorithm is sensitive to edge information. 
Some circular background areas and steel bars with blur edges are misidentified. 
In Liu $et.al.$\cite{kxiaohu2018research}, the contour-based algorithm relies on good contour extraction. 
It is sensitive to luminance variation of steel surface and edge blurring of steel bars.
Some single steel bars are misidentified multiple steel bars, and Some multiple steel bars are misidentified single steel bars.
The proposed method has better robustness for above disturbance, which performs well on environmental disturbance, luminance variation of steel surface and edge blurring of steel bars.
}
\label{fig:contrast}
\end{figure*}

\subsection{Results}
Three experiments are conducted in order to demonstrate the effectiveness of the proposed CNN-DC on automated steel bar counting and center localization. 
In the first experiment, the performance of CNN-DC on steel bar counting and center localization was analyzed. 
In the second experiment, the proposed CNN-DC was compared with other methods. 
In the third experiment, the analysis of two parameters (the test stride and the distance threshold $th_d$) in DC algorithm was conducted.

\subsubsection{The First Experiment}

The performance of CNN-DC on the steel bar dataset is shown in TABLE \ref{tab:ex1}. 
From TABLE \ref{tab:ex1}, it can be observed that CNN-DC can obtain high scores on Recall, Precision and F1, with average values of $0.9951$, $0.9976$ and $0.9963$, 
respectively, which indicates that CNN-DC can effectively identify center points on the steel bar images. 
Moreover, CNN-DC can achieve a high score on $Acc_r$ with an average value of $99.26\%$, 
which means that CNN-DC can have a good performance in steel bar counting. 
Moreover, the low $offset$ score ($4.1\%$) indicates that CNN-DC performs well on center localization. 
In addition, the calculation time of CNN-DC ($time=3.5862s$) indicates that the CNN-DC can meet the requirement of real-time processing for factory automation. 
The processes illustrating the application of CNN-DC on the steel bar dataset are shown in Fig. \ref{fig_flow}.

  \begin{figure*}
    \begin{tabular}{cc}
    \begin{minipage}{0.5\linewidth}
    \includegraphics[width=3.7in]{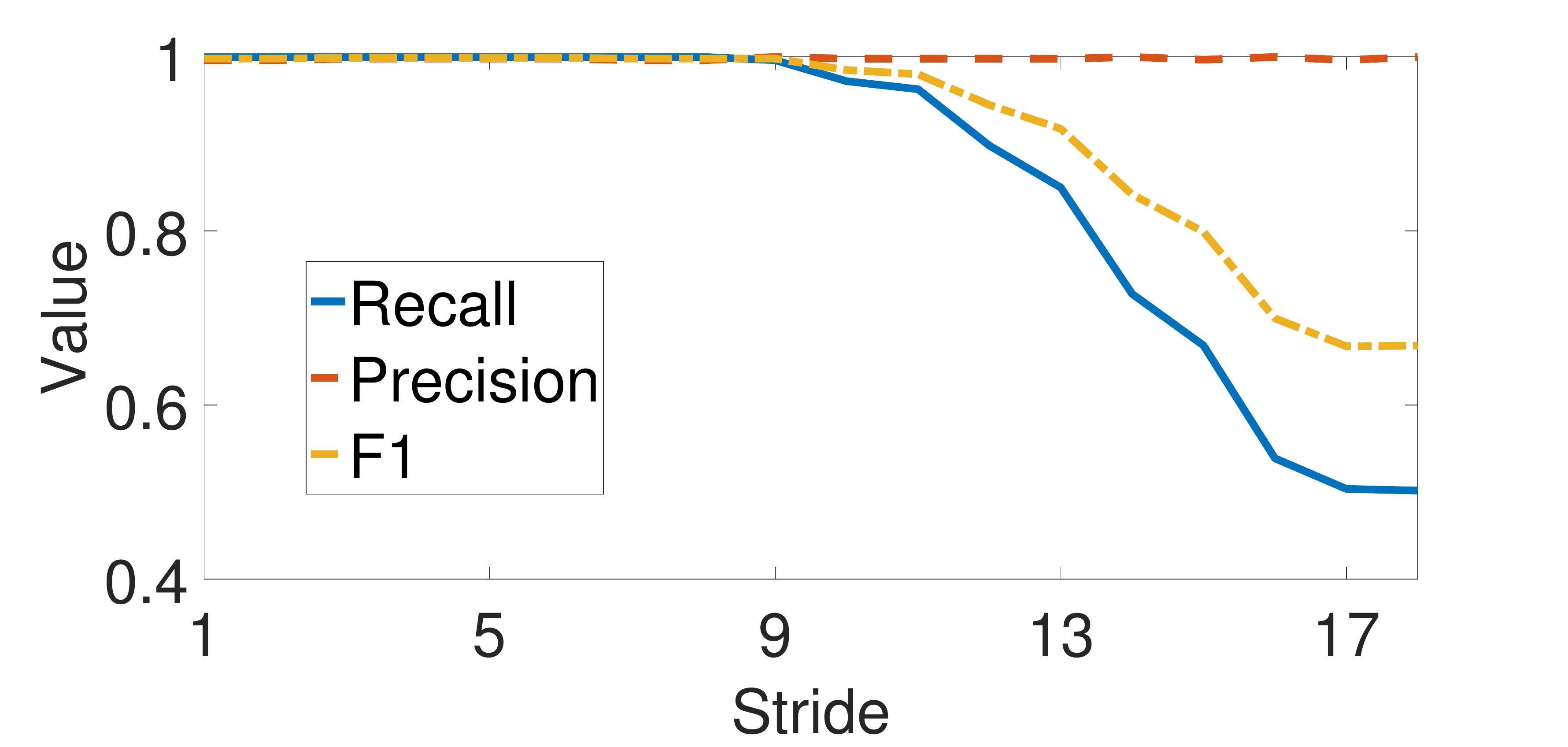}
    \end{minipage}
    \begin{minipage}{0.5\linewidth}
    \includegraphics[width=3.7in]{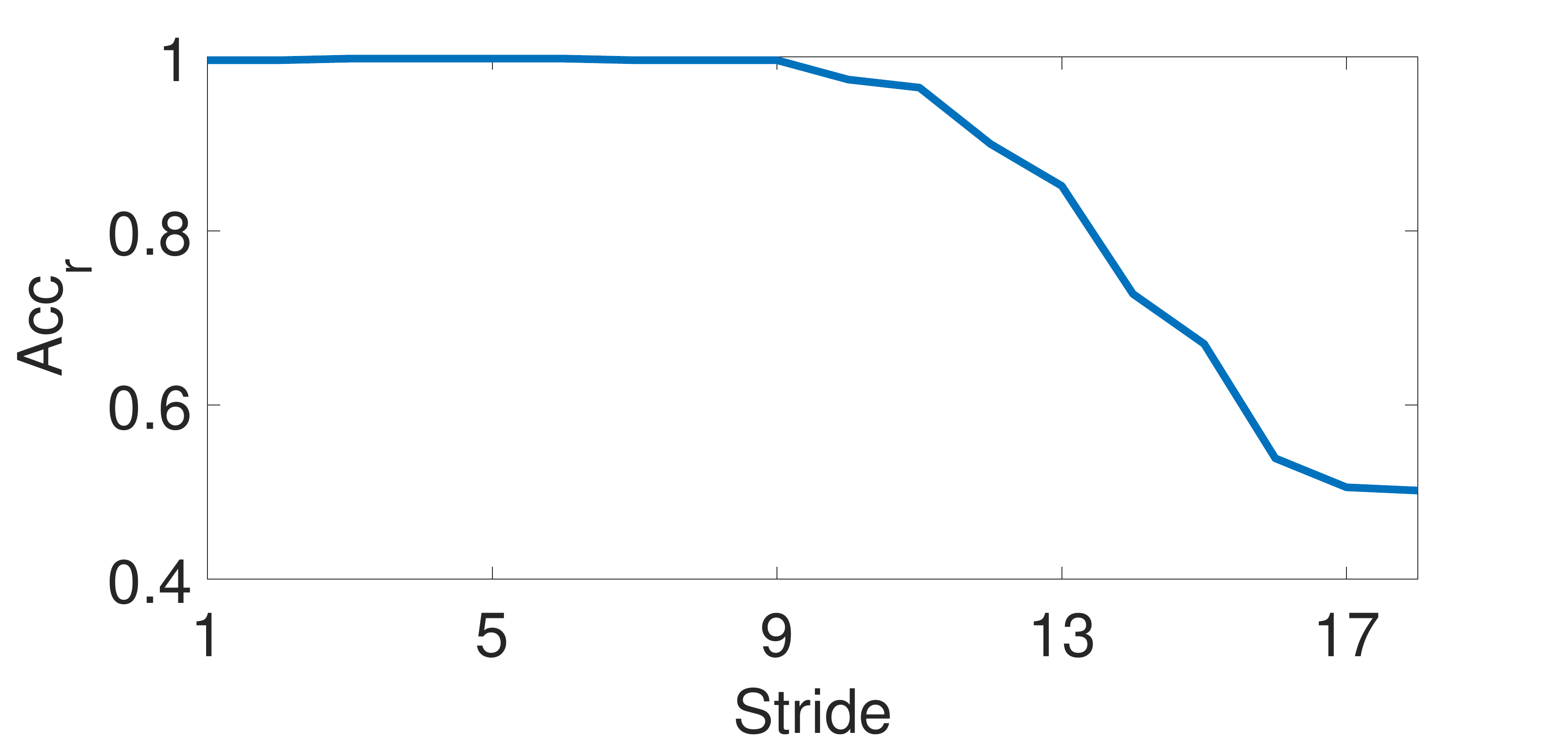}
    \end{minipage}
    \end{tabular}
    \center{\hspace{1.2cm}(a) \hspace{8.6cm} (b)}
    \begin{tabular}{cc}
      \begin{minipage}{0.5\linewidth}
      \includegraphics[width=3.7in]{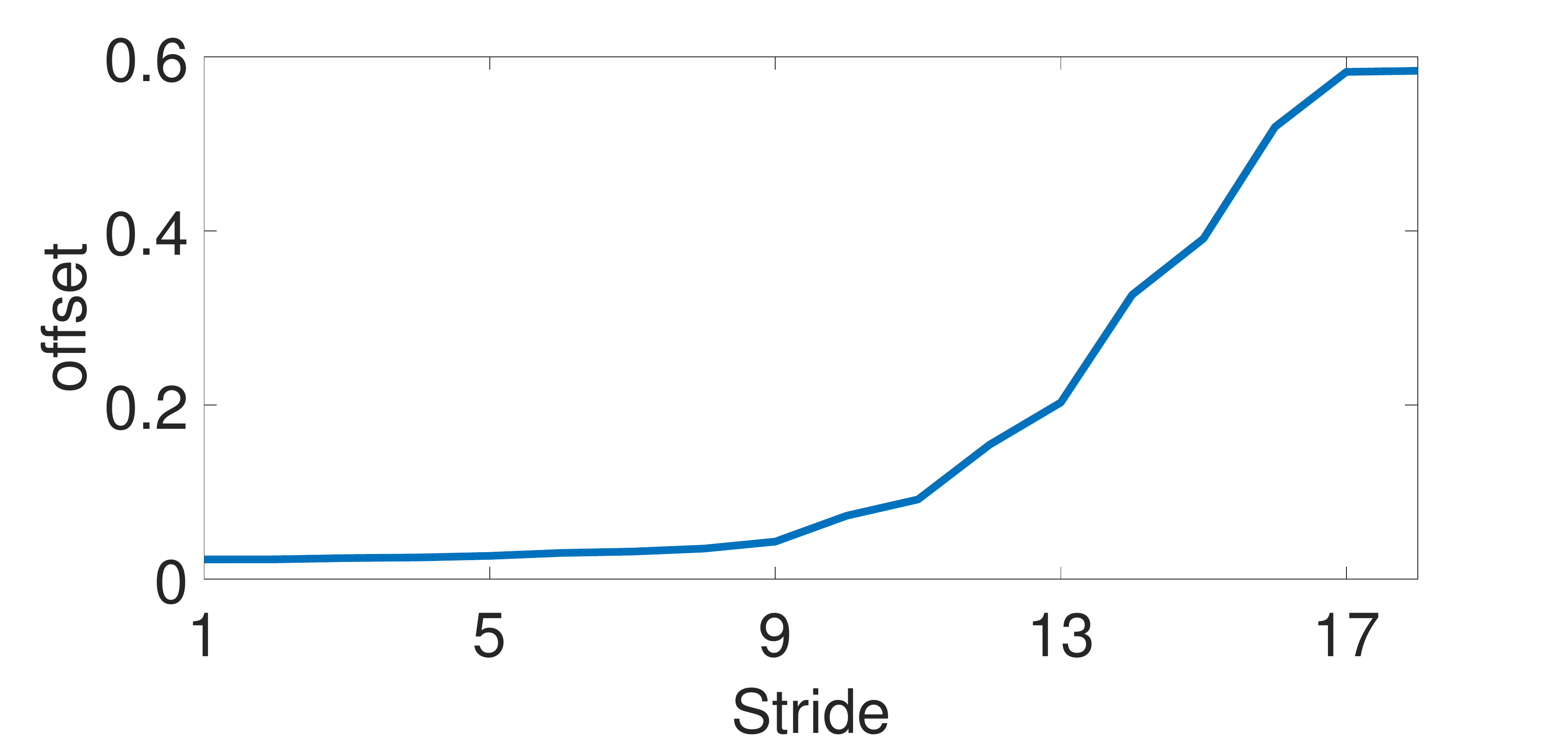}
      \end{minipage}
      \begin{minipage}{0.5\linewidth}
      \includegraphics[width=3.7in]{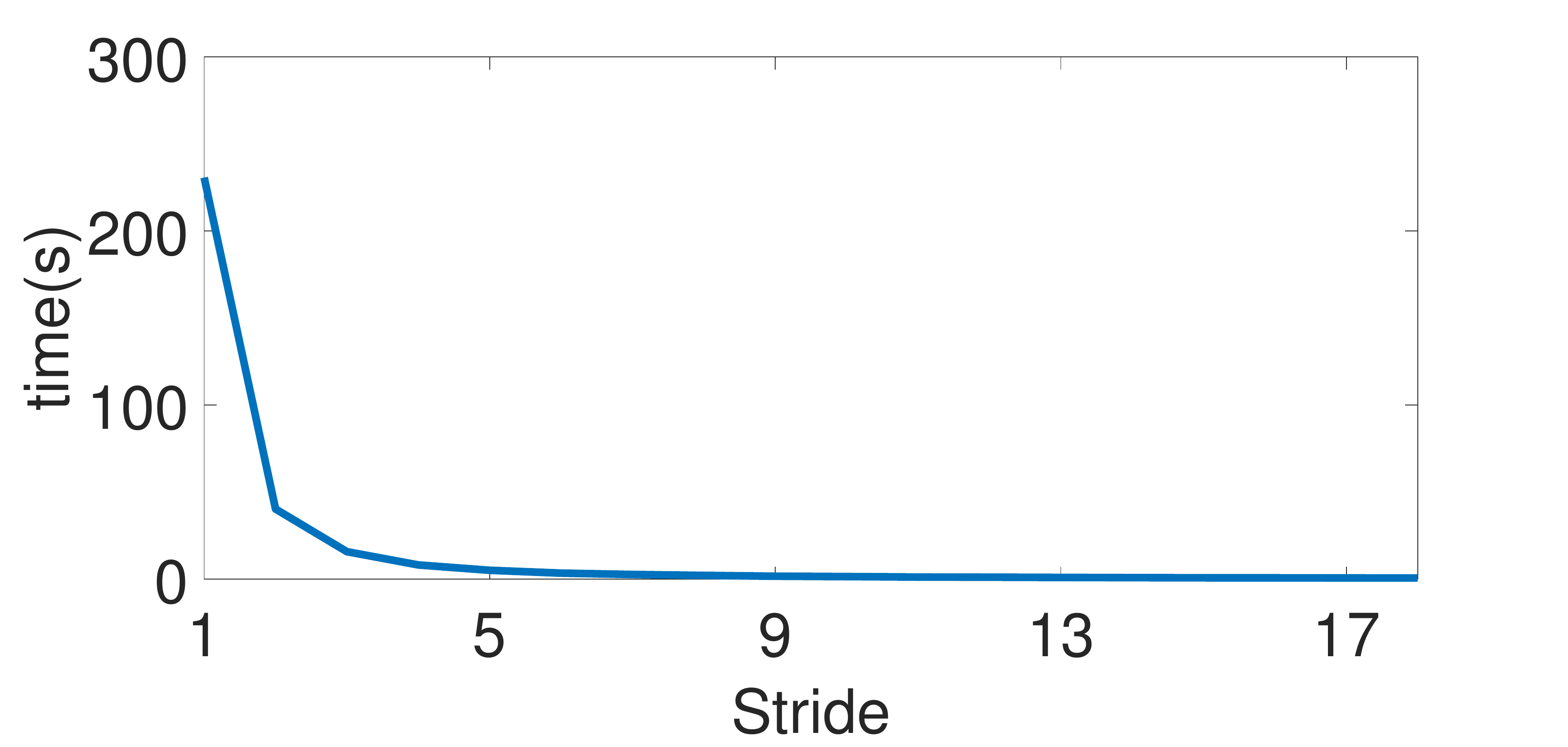}
      \end{minipage}
      \end{tabular}
      \center{\hspace{1.2cm}(c) \hspace{8.6cm} (d)}
  \caption{Variations of different evaluation metrics on the training data of the steel bar dataset with the increase of stride. (a) The variations of Recall, Precision and F1 with the increase of stride. (b) The variation of $Acc_r$ with the increase of stride. (c) The variation of offset with the increase of stride. (d) The variation of calculation time with the increase of stride. }
  \label{fig:stride1}
  \end{figure*}

  \begin{figure*}
    \begin{tabular}{cc}
    \begin{minipage}{0.5\linewidth}
    \includegraphics[width=3.7in]{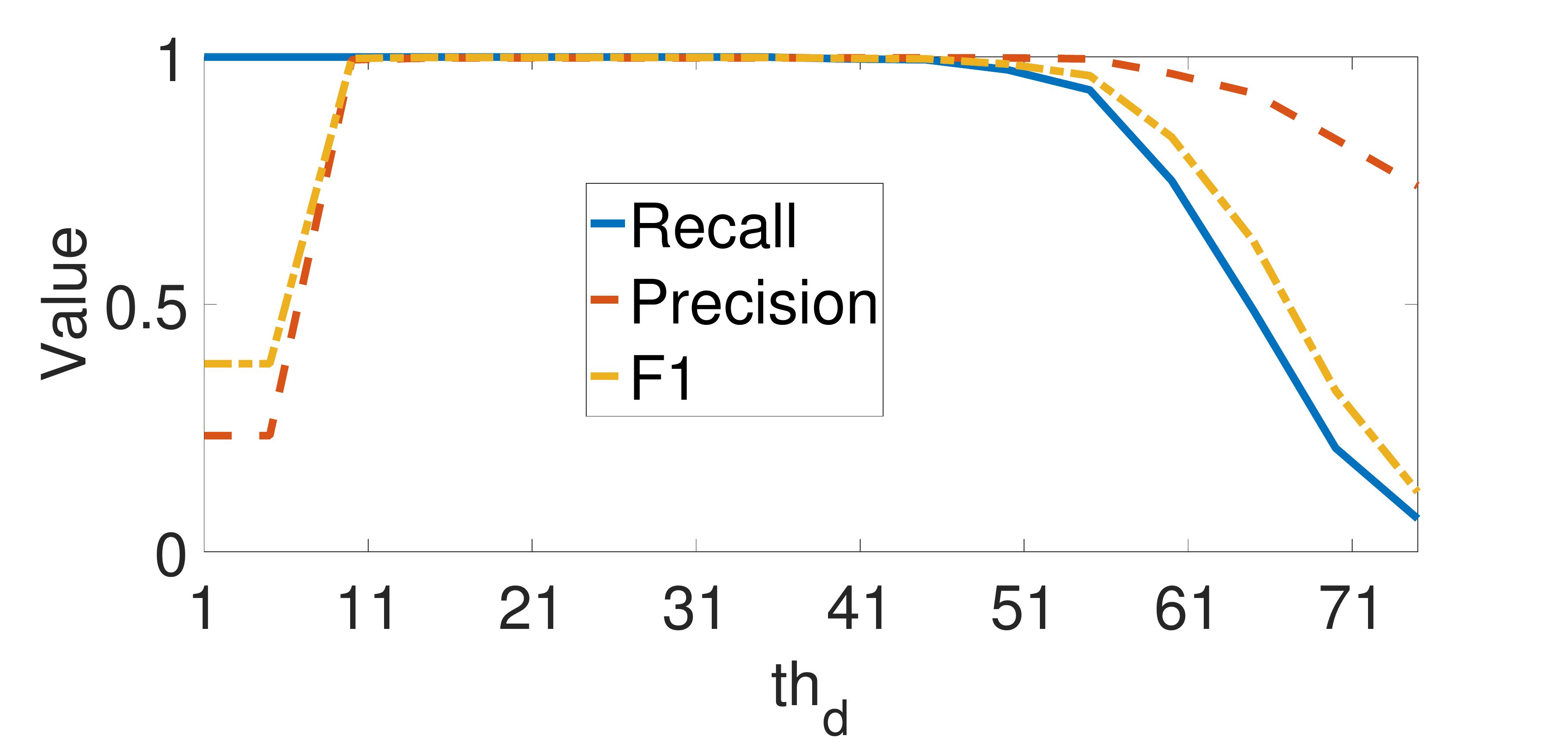}
    \end{minipage}
    \begin{minipage}{0.5\linewidth}
    \includegraphics[width=3.7in]{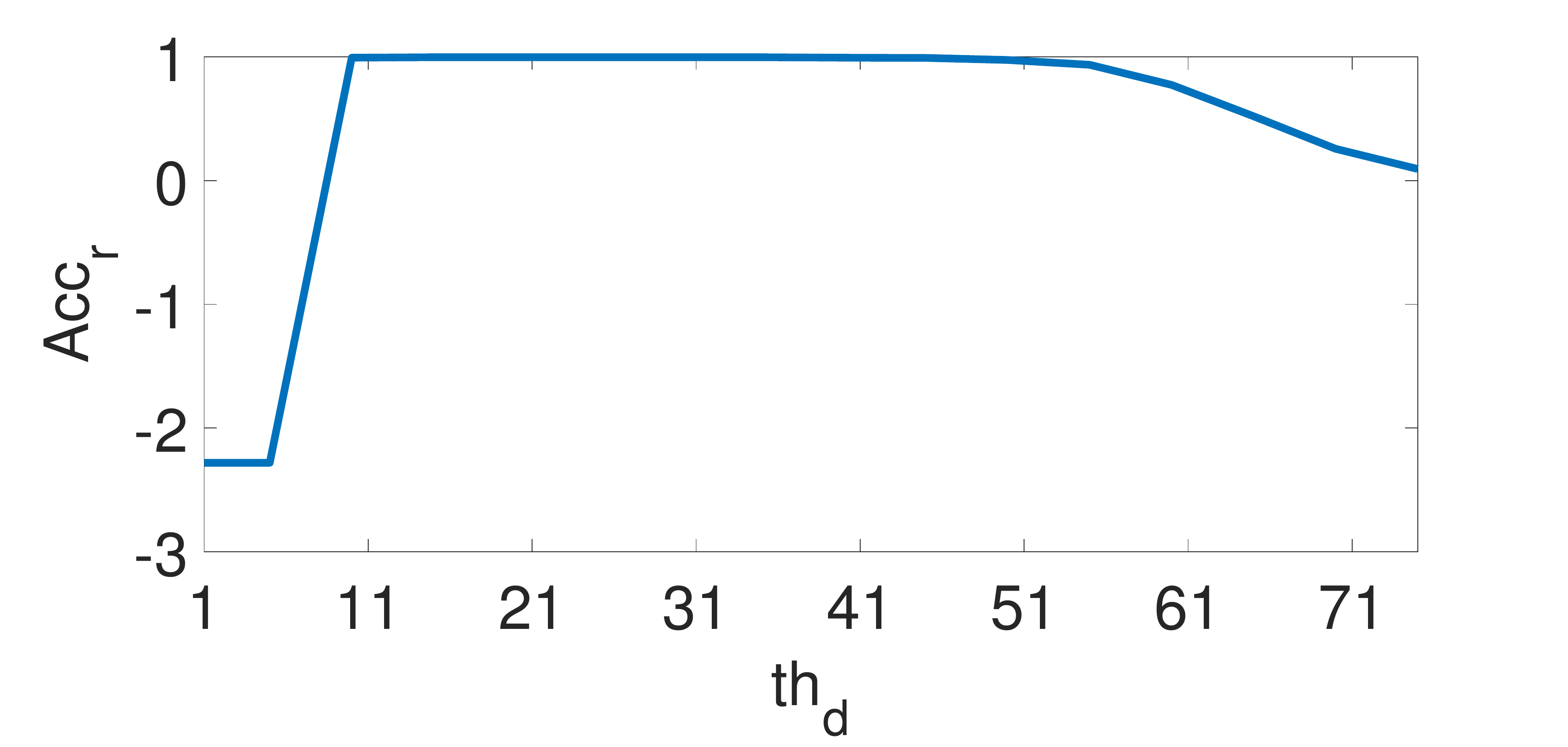}
    \end{minipage}
    \end{tabular}
    \center{\hspace{1.2cm}(a) \hspace{8.6cm} (b)}
    \begin{tabular}{cc}
      \begin{minipage}{0.5\linewidth}
      \includegraphics[width=3.7in]{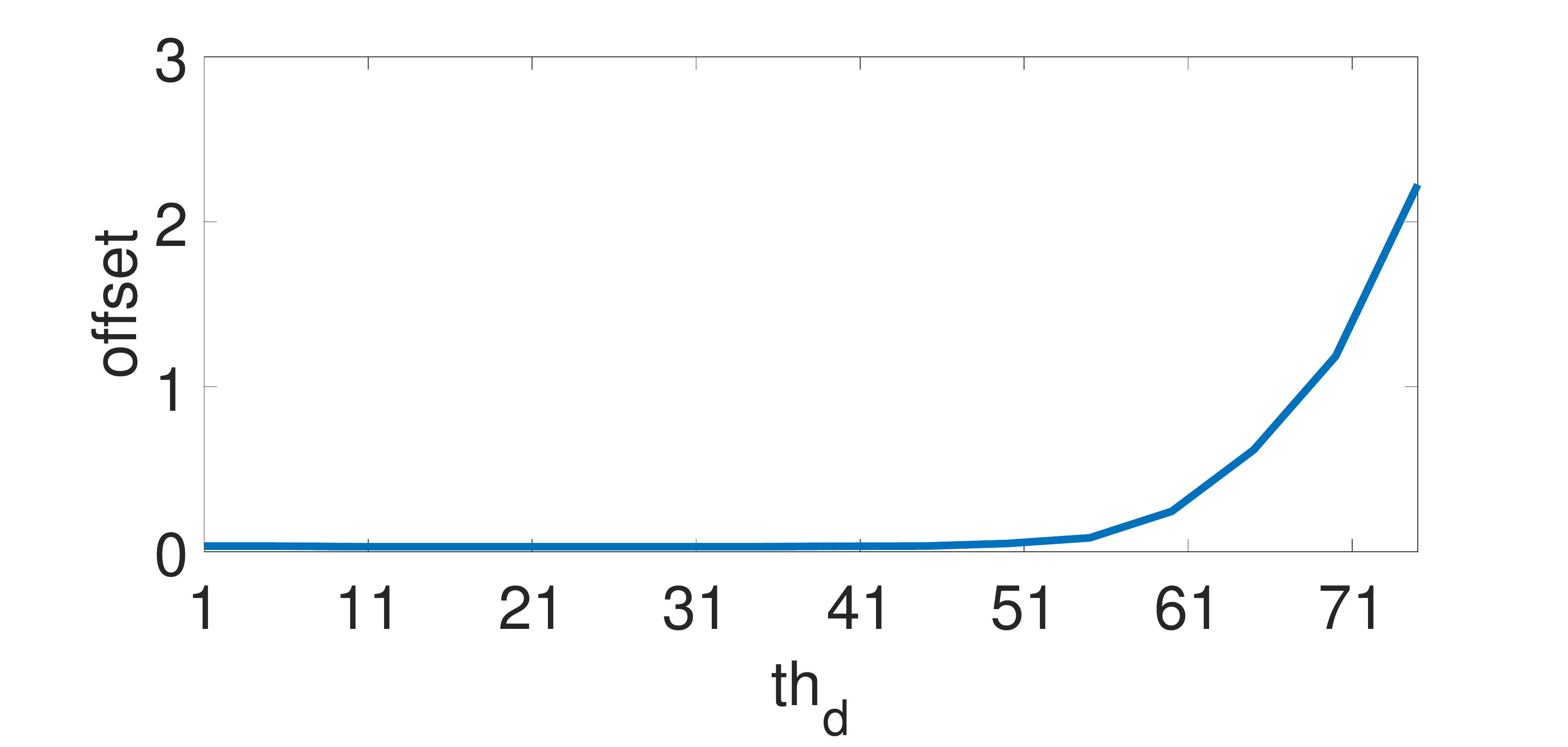}
      \end{minipage}
      \begin{minipage}{0.5\linewidth}
      \includegraphics[width=3.7in]{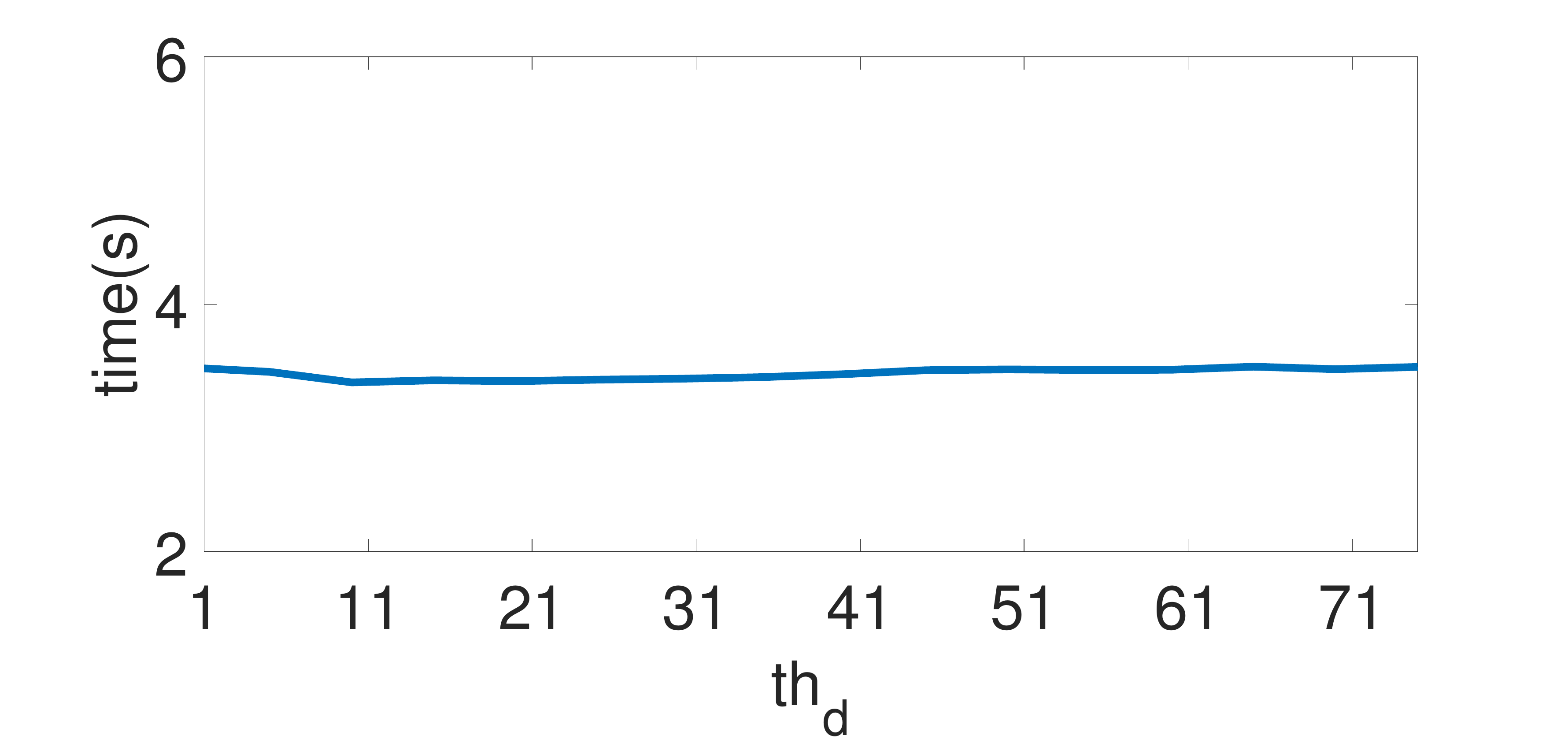}
      \end{minipage}
      \end{tabular}
      \center{\hspace{1.2cm}(c) \hspace{8.6cm} (d)}
  \caption{Variations of different evaluation metrics on the training data of the steel bar dataset with the increase of $th_d$. (a) The variations of Recall, Precision and F1 with the increase of $th_d$. (b) The variation of $Acc_r$ with the increase of $th_d$. (c) The variation of offset with the increase of $th_d$. (d) The variation of calculation time with the increase of $th_d$.}
  \label{fig:stride2}
  \end{figure*}

\begin{table*}
  \centering
  \caption{The comparison between CNN-DC and other methods}
    \begin{tabular}{ccccccc}
    \toprule
    \toprule
    Average & Recall & Precision & F1    & $Acc_r$ & $offset$ & times(s) \\
    \midrule
        Zhang $et.al.$\cite{gzhang2008bar} & 0.8864  & 0.9360  & 0.9103  & 94.69\%  & 15.83\% & 0.3023  \\
    Ying $et.al.$\cite{hying2010research} & 0.9617  & 0.8417  & 0.8975  & 85.68\%  & 12.58\% & 0.2404  \\
    Ghazali $et.al.$\cite{jghazali2017automatic} & 0.9778  & 0.9366  & 0.9566  & 95.56\%  & 10.30\% & 0.1346  \\
    Liu $et.al.$\cite{kxiaohu2018research} & 0.8123  & 0.6833  & 0.7420  & 80.99\%  & 25.58\% & \textbf{0.0313}  \\
    \textbf{Proposed} & \textbf{0.9951}  & \textbf{0.9976}  & \textbf{0.9963}  & \textbf{99.26\%}  & \textbf{4.11}\% & 3.5862  \\
    \bottomrule
    \bottomrule
    \end{tabular}%
  \label{tab:ex2}
\end{table*}%

\subsubsection{The Second Experiment}

In the first experiment, it has been demonstrated that CNN-DC performs well on steel bar counting and center localization. 
In order to further demonstrate the effectiveness of CNN-DC, we compare CNN-DC with other existing methods on the steel bar dataset. 
From TABLE \ref{tab:ex2}, it can be observed that CNN-DC outperforms other methods in terms of Recall, Precision, F1, $Acc_r$ and $offset$. 
As shown in Fig. \ref{fig:contrast}, the result obtained by other methods is sensitive to environmental disturbance, luminance variation of steel surface and edge blurring of steel bars.
The proposed method has better robustness for challenging environments.
Although the calculation time of CNN-DC is higher than other methods, it can be reduced by model acceleration methods.
With the help of network binarization\cite{rastegari2016xnor}, structured pruning\cite{luo2017thinet} and matrix decomposition\cite{kim2015compression},
the calculation time can be further reduced and will be the next step in our future work.


\subsubsection{The Third Experiment}
In our work, $stride=6$ and $th_d=20$ are used, which were selected based on the experiments on the training data of the steel bar dataset. 
In order to illustrate the reasons of choosing these values, 
the variations of the evaluation metrics: Recall, Precision, F1, $offset$ and the calculation time on the training data of the steel bar dataset by varying $stride$ and $th_d$ are given in Fig. \ref{fig:stride1} and \ref{fig:stride2}. 
From Fig. \ref{fig:stride1}, it can be observed that CNN-DC can have a good performance on the training data of the steel bar dataset as $stride$ varies in [1,9]. 
In the meantime, CNN-DC can have a low calculation time  as $stride$ varies in [5,18]. 
From the above observation, CNN-DC performs overall better as $stride$ varies in [5,9], and $stride=6$ is chosen randomly from [5,9]. 
From Fig. \ref{fig:stride2}, it can be observed that CNN-DC can maintain high scores of Recall, Precision, F1 and $Acc_r$ as $th_d$ varies in [11,51]. 
CNN-DC can also have a low $offset$ value as $th_d$ varies in [1,51]. 
Moreover, the calculation time varies very slightly with change of $th_d$. 
From the above observation, CNN-DC can obtain a good performance as $th_d$ varies [11,51], and $th_d=20$ is selected randomly from [11,51].


\section{Conclusions}
Automated steel bar counting and center localization are of great significance in factory automation of steel bars. 
Steel bar counting and center localization are traditionally performed by skilled workers, which are tedious and time-consuming. 
In order to alleviate the burdens of workers on steel bar counting and center localization, 
an effective framework called CNN-DC is proposed to achieve steel bar counting and center localization simultaneously. 
The proposed CNN-DC framework first performs candidate center point detection with a deep convolutional neural network, 
which is followed by a Distance Clustering algorithm to cluster the candidate center points and obtain the center locations of steel bars. 
The experimental results show that the proposed CNN-DC framework performs well on steel bar counting and center localization, 
achieving $Recall=0.9951$, $Precision=0.9976$, $F1=0.9963$, $Acc_r=99.26\%$, $offset=4.11\%$ and $time=3.5862s$.

In the future, more steel bar images will be collected by negotiating with the managers of steel bar factories in order to further validate the effectiveness of CNN-DC. 
Applying CNN-DC real time in steel bar factories is also planned in order to validate the practicability of the proposed method.

\bibliography{bare_jrnl}
\bibliographystyle{IEEEtran}
%


\end{document}